\documentclass[final,3p,times,twocolumn,authoryear]{elsarticle}

\usepackage{amssymb}
\usepackage{amsmath}
\usepackage{lineno}

\usepackage{orcidlink}
\usepackage{tabularx}
\usepackage[ruled,linesnumbered]{algorithm2e}
\usepackage{latexsym}
\usepackage{xcolor}
\usepackage{subfigure}
\usepackage{multirow}
\usepackage{multicol}
\usepackage{float}
\usepackage{makecell}
\usepackage{booktabs}
\usepackage{threeparttable}
\usepackage{enumitem}
\usepackage[ruled,linesnumbered]{algorithm2e}
\usepackage{cancel}
\usepackage{ulem}
\newcolumntype{"}{!{\hskip\tabcolsep\vrule width 1.5pt\hskip\tabcolsep}}
\usepackage {soul}

\soulregister{\cite}7
\soulregister{\sout}7
\soulregister{\ref}7
\soulregister{\it}7
\soulregister{\myeq}7

\usepackage{hyperref}
\hypersetup{
    colorlinks=true,
    linkcolor=blue,
    filecolor=blue,      
    urlcolor=blue,
    anchorcolor=blue,
    citecolor=blue,
}
\usepackage{color}

\usepackage[]{changes}
\definechangesauthor[name={zou}, color=red]{zou}

\definechangesauthor[name={sxf}, color=green]{sxf}

\newcommand{\myeq}[1]{Eq.~(\ref{#1})}
\begin{document}
\begin{frontmatter}

\title{GIMS: Image Matching System Based on Adaptive Graph Construction and Graph Neural Network}

\author[label1]{Xianfeng Song}
\ead{misongxf@mail.scut.edu.cn}
\author[label1]{Yi Zou\corref{cor1}}
\ead{zouyi@scut.edu.cn}
\author[label1]{Zheng Shi}
\ead{mieesz@mail.scut.edu.cn}
\author[label4]{Zheng Liu}
\ead{zheng.liu@tuta.io}

\cortext[cor1]{Corresponding author.}
\affiliation[label1]{organization={School of Microelectronics, South China University of Technology},
            city={Guangdong},
            country={China}}
\affiliation[label4]{organization={Faculty of Applied Science, University of British Columbia},
            city={Kelowna},
            country={Canada}}

\begin{abstract}
Feature-based image matching has extensive applications in computer vision. Keypoints detected in images can be naturally represented as graph structures, and Graph Neural Networks (GNNs) have been shown to outperform traditional deep learning techniques. Consequently, the paradigm of image matching via GNNs has gained significant prominence in recent academic research. In this paper, we first introduce an innovative adaptive graph construction method that utilizes a filtering mechanism based on distance and dynamic threshold similarity. This method dynamically adjusts the criteria for incorporating new vertices based on the characteristics of existing vertices, allowing for the construction of more precise and robust graph structures while avoiding redundancy. We further combine the vertex processing capabilities of GNNs with the global awareness capabilities of Transformers to enhance the model's representation of spatial and feature information within graph structures. This hybrid model provides a deeper understanding of the interrelationships between vertices and their contributions to the matching process. Additionally, we employ the Sinkhorn algorithm to iteratively solve for optimal matching results. Finally, we validate our system using extensive image datasets and conduct comprehensive comparative experiments. Experimental results demonstrate that our system achieves an average improvement of 3.8x-40.3x in overall matching performance. Additionally, the number of vertices and edges significantly impacts training efficiency and memory usage; therefore, we employ multi-GPU technology to accelerate the training process. Our code is available at \href{https://github.com/songxf1024/GIMS}{https://github.com/songxf1024/GIMS}.

\end{abstract}



\end{frontmatter}

\section{Introduction}
\label{sec:introduction}
As a cornerstone research area in computer vision, image matching has numerous applications, including object detection~\citep{ma2015robust, rashid2019object}, image stitching~\citep{ravi2020development, sharma2020image}, Structure-from-Motion (SfM)~\citep{schonberger2016structure}, visual localization~\citep{sattler2018benchmarking, taira2018inloc}, and pose estimation~\citep{grabner20183d, persson2018lambda}. Image matching can be categorized into traditional methods, deep learning-based methods, and hybrid methods. Traditional methods typically rely on the detection and matching of keypoints in images, such as SIFT~\citep{lowe1999object}, SURF~\citep{bay2006surf}, and ORB~\citep{rublee2011orb}. These techniques achieve image matching by identifying and comparing keypoints, which are inherently designed to resist changes in image scaling, rotation, and partial occlusion. However, these methods may be ineffective in dealing with complex image variations, such as drastic lighting changes or significant shifts in perspective. Moreover, each keypoint contains only its own feature and does not utilize features from neighboring keypoints. 

In contrast, deep learning methods~\citep{tian2020hynet, 9852191, 9591230} enhance the robustness of image matching by training neural networks to learn deep feature representations of images. These methods typically employ Convolutional Neural Networks (CNNs) or Transformers to extract image features and learn matching patterns directly from data through an end-to-end process. Deep learning methods excel at handling non-linear image variations and complex patterns, thereby delivering superior performance. However, similar to traditional methods, deep learning methods often capture either local or global features and fail to effectively integrate both. 

Hybrid methods~\citep{barroso2019key, chen2023shape, rodriguez2019sift} incorporate the advantages of both traditional and deep learning methods. These methods aim to improve the accuracy and robustness of image matching by either integrating handcrafted features into a deep learning framework or combining them at the feature extraction stage. The work~\citep{song2023carhynet} successfully explored the combination of handcrafted and deep features for matching at the decision level and achieved excellent results.

Additionally, we recognize that the aforementioned methods often overlook the interdependencies among keypoints, such as positional relationships. We recognize the opportunity to approach image matching through an alternative paradigm. Keypoints extracted by handcrafted features can form graph structures, prompting us to consider whether image matching can be studied in a different way. However, traditional CNNs have difficulty handling such irregular data. Fortunately, unlike highly structured data such as images and text, graphs, which are composed of vertices and edges, excel at representing and analyzing data in non-Euclidean spaces. Moreover, Graph Neural Networks (GNNs), designed specifically for graph data, can directly process graph structures and are considered crucial for advancing artificial intelligence from ``perceptive intelligence" to ``cognitive intelligence". GNNs can learn universal paradigms for any graph structure, and any improvements can be generalized across various fields, thus having wide applications. In fact, GNNs have shown significant potential in the field of image matching in recent years. In order to make images suitable for GNN processing, various methods are available to construct graphs. However, these methods often result in graphs with an excessive number of vertices or edges and contain isolated vertices or subgraphs.

To address these issues, we propose a new image matching system based on two novel methods that work synergistically. First, we employ a similarity-based adaptive graph construction method to minimize vertex and edge redundancy by selectively creating edges between highly similar vertex pairs. Second, we leverage the merits of both GNN and Transformer to integrate local structure with global information for robust image matching. We summarize the main contributions of the proposed system as follows.


\begin{itemize}
    \item \textbf{To effectively reduce graph redundancy, we introduce a new similarity-based adaptive graph construction method.} By dynamically adjusting the graph construction according to the similarity between feature descriptors, we add edges only between vertex pairs with high similarity. This data-driven method means the graph construction process is directly determined by the characteristics and interrelationships of the data, thereby better capturing and utilizing the intrinsic structure and patterns within image data, as well as efficiently controlling the graph density. 

    \item \textbf{To successfully integrate local structure with global information, we propose a novel method that combines GNN with Transformer.} First, GNN aggregates information from neighboring vertices on the graph to update each vertex, capturing complex relationships among local structures. The Transformer then captures long-distance dependencies. The combination of these methods effectively fuses local graph structure and global features.

    \item \textbf{To offer a holistic view of the proposed approach, we provide comprehensive comparative experiments of the proposed system and existing methods.} We compare classical methods, state-of-the-art methods, and the proposed method on standard large-scale benchmark image datasets. Additionally, we explore performance in different scenarios, thus comprehensively evaluating the effectiveness and applicability of each method.

    \item \textbf{To improve model training efficiency, we employ a multi-GPU parallel acceleration technique.} Compared to traditional handcrafted feature methods, deep learning models, especially GNN and Transformer, typically require large datasets for training. To efficiently reduce training time, we implement a data parallelism strategy to accelerate the training process. 
\end{itemize}

As we show later in Section \ref{sec:experiments}, the proposed graph-based image matching system based on these novel methods significantly improves image matching performance. This paper is organized as follows. We analyze the related work in Section \ref{sec:relwork}. In Section \ref{sec:method}, we describe the proposed graph construction and graph matching method in detail. We describe the experimental setup and comparative studies in Section \ref{sec:experiments}. We also share our thoughts on the limitations of the current work and areas for future exploration in Section \ref{sec:discussion}. Finally, we conclude this paper in Section \ref{sec:conclusion}.

\begin{figure*}[!ht]
    \centering
    \includegraphics[width=0.92\linewidth]{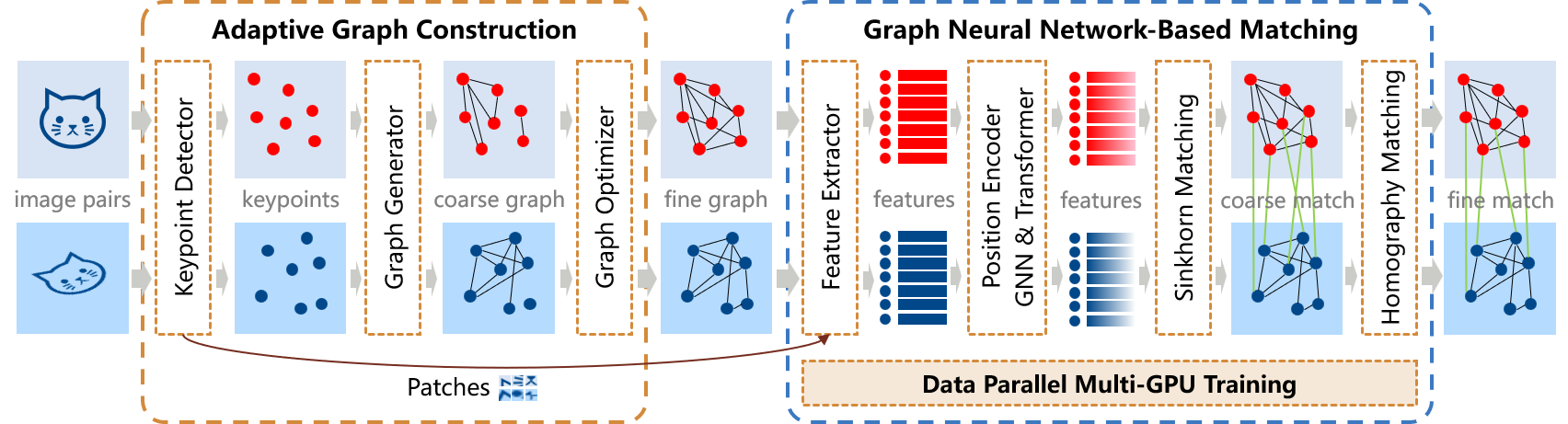}
    \caption{\textbf{Overall system architecture.} The architecture consists of two parts: \textit{Adaptive Graph Construction} (AGC) and \textit{Graph Neural Network-Based Matching} (GM). After inputting image pairs, the first part converts images to graphs, and then the second part performs graph matching. Note that features of vertices are generated from patches constructed from keypoints. Finally, the system outputs the matching results. Moreover, to effectively reduce training time, we use \textit{Data Parallel Multi-GPU Training} to accelerate model training.}
    \label{fig:overall_architecture}
    \vspace{-1.5em}
\end{figure*}

\section{Related Work}
\label{sec:relwork}

\subsection{Conventional Image Matching}
Image matching has a long research history and still attracts much attention today. In early research, it primarily relied on handcrafted features and matching methods. The most commonly used methods, such as SIFT (Scale-Invariant Feature Transformation)~\citep{lowe1999object}, SURF (Speeded-Up Robust Features)~\citep{bay2006surf}, and ORB (Oriented FAST and Rotated BRIEF)~\citep{rublee2011orb}, detect keypoints within images and generate a distinct feature descriptor for each keypoint, achieving invariance to changes in lighting, rotation, and scale. These methods are often accompanied by clear mathematical explanations and high efficiency. However, they may lack robustness when dealing with highly complex image variations, such as extreme changes in perspective, blurring, and occlusion. Besides, hashing-based methods~\cite{10.5555/3060832.3060860, cao2017hashnet} enable efficient similarity computation, but they are often insufficient for tasks requiring pixel-level alignment or precise spatial correspondence.

With the significant advancement of CNNs in image processing, innovative methods based on deep learning have been applied to improve the accuracy and robustness of image matching. For keypoint detection, MagicPoint~\citep{detone2017toward} is a self-supervised corner detection algorithm using a CNN and is capable of identifying visually significant keypoints. SuperPoint~\citep{detone2018superpoint} extends MagicPoint to provide an end-to-end framework for keypoint detection and descriptor generation via a fully convolutional network, optimizing the efficiency and effectiveness of the entire process. Additionally, researches such as HardNet~\citep{mishchuk2017working}, SOSNet~\citep{tian2019sosnet}, and HyNet~\citep{tian2020hynet} focus on the learning of feature descriptors. For these works, the detection of keypoints can be done using various methods, such as SIFT or MagicPoint. DeDoDe~\citep{edstedt2024dedode} decouples descriptor and detector learning while having aligned objectives, demonstrating deep learning advances in the field of image matching. Furthermore, there are researchers dedicated to achieving end-to-end image matching~\citep{9852191}.

In recent years, considering the distinct advantages of traditional and deep learning methods, researchers have begun to explore how to combine handcrafted features with deep features. Key.Net~\citep{barroso2022key} integrates handcrafted features into the shallow layers of a convolutional network and uses them as anchor structures for subsequent convolutional layers. SIFT-AID~\citep{rodriguez2019sift} uses SIFT as the first stage to generate keypoints and then uses CNN to generate affine invariant descriptors. CAR-HyNet~\citep{song2023carhynet} then fuses handcrafted features and deep features from a decision-level perspective.

\subsection{Graph-Based Image Matching}
However, the aforementioned methods can only process highly structured data, overlooking potential interdependencies between keypoints. A graph is a data structure composed of vertices and edges, which can capture complex relationships effectively between irregular data in non-Euclidean space. The success of GNNs in various fields has increased interest in their application to image matching.

\subsubsection{\textbf{Graph Construction}}
There are many classic works on converting images to graphs. Delaunay triangulation~\citep{delaunay6sphere} forms triangles from discrete vertices, ensuring no four vertices are co-circular. This method naturally constructs graph structures considering spatial proximity, but is mainly applicable for data with obvious spatial relationships. Another work is to construct a \textit{k}-Nearest Neighbors graph~\citep{dong2011efficient}, which is constructed by connecting each vertex to the \textit{k} vertices that are most similar to it based on a specific similarity measure. In addition, the Minimum Spanning Tree (MST)~\citep{prim1957shortest} generates a tree with the minimum sum of edge weights among all vertices of the graph, ensuring that no cycles are formed. It can be used to connect vertices to minimize the total connection distance. The Epsilon-Neighborhood~\citep{ester1996density} method connects all vertex pairs within a fixed distance $\epsilon$. This method relies on a predetermined distance threshold to determine which vertices should be connected to each other. However, it also results in some vertices that are farther apart not being connected.

\subsubsection{\textbf{Graph Matching}}
SuperGlue~\citep{sarlin2020superglue} treats keypoints as vertices in a graph and constructs a graph using descriptors and positional information of keypoints. These keypoints in two graphs are then processed using a bidirectional attention mechanism to perform the vertex messaging and iterative update process. LightGlue~\citep{lindenberger2023lightglue} enhances SuperGlue by introducing a method to dynamically adjust the depth and width of the network according to the difficulty of matching tasks, achieving speed improvements of 4 to 10 times. The above works leverage the characteristics of Transformers to indirectly implement the vertex update process of GNNs, and each update involves all vertices. KeyGNN~\citep{jiang2023improving} proposes a structure-aware sparse graph neural network, introducing an Informative Keypoint Exploration (IKE) module and a Guided Sparse attention (GS) module to achieve a good balance between accuracy and efficiency. However, its performance is sensitive to the quality of the initial seed matches. MGMN~\citep{9516695} combines cross-level node-graph matching and global graph-graph interactions to compute the graph similarity between any pair of graphs end-to-end. SAT~\citep{chen2022structure} explicitly incorporates the graph structure to capture structural interactions between vertices, exhibiting superior performance on multiple graph prediction benchmarks and offering greater interpretability compared to other methods. The recent OmniGlue~\citep{jiang2024Omniglue} integrates SuperPoint for keypoint detection, uses DINOv2 to generate features, and employs SuperGlue for matching, leading to a significant increase in its demand for memory and computational resources. However, the computational demands of GNNs are related to the number of vertices and edges. Therefore, it is necessary to explore strategies to minimize the number of vertices and edges without significantly compromising performance.

\section{The Proposed Method}
\label{sec:method}
Incorporating positional information in GNNs helps the model to understand the physical relationships between vertices. Introducing GNNs into image matching is expected to significantly improve the accuracy and efficiency of matching. In this paper, we propose a novel image matching system which consists of \textit{Adaptive Graph Construction} (AGC) and \textit{Graph Neural Network-based Matching} (GM), and the overall architecture is shown in Fig.~\ref{fig:overall_architecture}. The common notations within the paper are summarized in Table~\ref{tab:notations}.

\subsection{Adaptive Graph Construction}
\begin{figure}[!htb]
    \vspace{-0.5em}
    \centering
    \setlength{\abovecaptionskip}{0em}
    \setlength{\belowcaptionskip}{0em}
    \includegraphics[width=\linewidth]{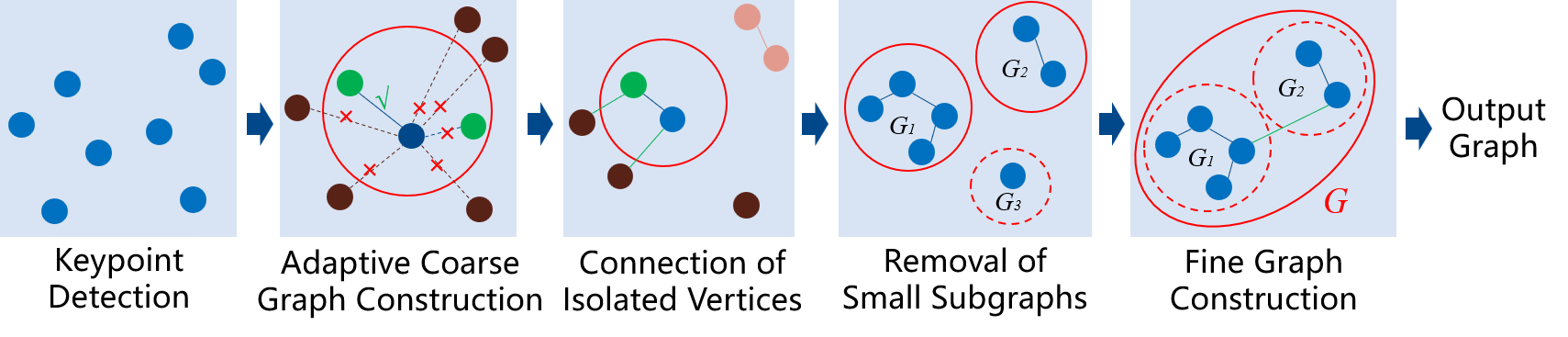}
    \caption{\textbf{Overall flow of adaptive graph construction}. This flow consists of five steps and aims to avoid redundant vertices and edges as much as possible while maintaining a robust graph structure.}
    \label{fig:graph_construction_overall_flow}
    \vspace{-0.5em}
\end{figure}

The graph construction method is critical. An inappropriate graph construction method can cause many issues, such as numerous isolated subgraphs, excessive vertices and edges. These issues seriously affect the learning efficiency and performance of GNNs. We adopt a coarse-to-fine strategy by first constructing an adaptive coarse graph based on distance and similarity awareness, and then adjusting vertices and subgraphs to form a fine graph, The overall flow is shown in Fig.~\ref{fig:graph_construction_overall_flow} and Algorithm~\ref{alg:agc}. The proposed graph construction method avoids redundant vertices and edges while maintaining a robust graph structure. Finally, we compare several graph construction methods to show its superiority.

\begin{algorithm}[!h]
    \scriptsize 
    \SetAlgoNlRelativeSize{-2} 
    \SetAlgoSkip{smallskip} 
    \SetAlgoInsideSkip{smallskip} 
    \SetInd{0.5em}{0.8em} 
    \caption{Adaptive Graph Construction}
    \label{alg:agc}
    \KwIn{Input image $I$, distance threshold $\beta$, percentile threshold $\alpha$, minimum subgraph size $\theta$}
    \KwOut{A constructed graph $G = (V, E)$}

    \tcp{Step 1: Keypoint Detection}
    $\{v_i\}_{i=1}^n \gets \mathrm{SIFT}(I)$ \tcp*{detect $n$ keypoints} 
    $\text{position } p_i,\ \text{score } s_i,\ \text{scale } \sigma_i \gets v_i$ \tcp*{provided by SIFT}
    $\mathcal{G} \gets \mathrm{GaussianPyramid}(I)$ \tcp*{rebuild multi-scale pyramid}
    $\mathcal{P}_i \gets \mathrm{Crop}(\mathcal{G}_{\sigma_i}, p_i, 32 \times 32)$ \tcp*{extract patches from $\mathcal{G}_{\sigma_i}$}
    $\mathbf{f}_i \gets \mathrm{CAR\text{-}HyNet}(\mathcal{P}_i)$ \tcp*{compute deep descriptors}

    \tcp{Step 2: Adaptive Coarse Graph Construction}
    $\mathcal{T} \gets \mathrm{KDTree}(\{p_i\})$ \tcp*{build KDTree for spatial filtering}
    $\mathcal{C} \gets \mathrm{PairsWithinRadius}(\mathcal{T}, \beta)$ \tcp*{get candidate edges via query\_pairs}
    $C_{ij} \gets \cos(\mathbf{f}_i, \mathbf{f}_j)\ \forall (i,j) \in \mathcal{C}$ \tcp*{compute cosine similarities}
    $\gamma \gets \mathrm{Percentile}(\{C_{ij}\}, \alpha)$ \tcp*{adaptive similarity threshold}
    $E \gets \{(i,j) \in \mathcal{C} \mid C_{ij} \geq \gamma\}$,\quad $V \gets \{v_i\}$

    \tcp{Step 3: Connection of Isolated Vertices}
    \ForEach{$v_i \in V$ with $\deg(v_i) = 0$}{
        $v_j \gets \mathrm{NearestNeighbor}(\mathcal{T}, v_i)$ \tcp*{find nearest neighbor using KDTree}
        $E \gets E \cup \{e_{ij}\}$ \tcp*{connect isolated vertex}
    }

    \tcp{Step 4: Removal of Small Subgraphs}
    Let $\{G_k = (V_k, E_k)\}$ be the set of connected components of $G$\;
    \ForEach{$G_k$}{
        \If{$|V_k| < \theta$}{
            remove $G_k$ from $G$ \tcp*{filter small subgraph}
        }
    }

    \tcp{Step 5: Fine Graph Construction}
    \While{$G$ has more than one connected component}{
        $\tilde{p}_i \gets \frac{1}{|V_i|} \sum_{v_j \in V_i} p_j$ for each $G_i$ \tcp*{compute centroid of each subgraph}
        Find closest pair $(\tilde{p}_i, \tilde{p}_j)$ using KDTree \tcp*{find closest two subgraphs}
        $(v_u, v_v) \gets \arg\min_{v_u \in V_i,\, v_v \in V_j} \|p_u - p_v\|$ \tcp*{find closest node pair across components}
        $E \gets E \cup \{e_{uv}\}$ \tcp*{connect the two components}
    }

    \Return{Final constructed graph $G = (V, E)$}
\end{algorithm}

\begin{table}[]
    \centering
    \caption{\textbf{Notations}.}
    \setlength{\abovecaptionskip}{0em}
    \setlength{\belowcaptionskip}{0em}
    \label{tab:notations}
    \resizebox{0.8\columnwidth}{!}{%
        \renewcommand{\arraystretch}{1.4}
        \begin{tabular}{|c||l|}
            \hline
            $G$ & The constructed graph \\ \hline
            $V, E$ & The vertex set and the edge set of graph $G$ \\ \hline
            $|V|, |E|$ & The order and size of graph $G$ \\ \hline
            $A, B$ & Two images to be matched \\ \hline
            $m, n$ & The total number of vertices in images $A$ and $B$ \\ \hline
            $N(i)$ & The neighbor set of vertex $v_i$ \\ \hline
            $G_i$ & The $i$-th subgraph of graph $G$ \\ \hline
            $v_i$ & The $i$-th vertex \\ \hline
            $e_{ij}$ & The edge between vertices $v_i$ and $v_j$ \\ \hline
            $\mathbf{C}$ & The upper triangular matrix of cosine similarities \\ \hline
            $\tilde{C}$ & The set of flattened and sorted matrix $\mathbf{C}$ \\ \hline
            $c_{ij}$ & The cosine similarity of vertices $v_i$ and $v_j$ \\ \hline
            $\alpha$ & The percentile threshold in adaptive graph construction \\ \hline
            $\gamma$ & The similarity threshold in adaptive graph construction \\ \hline
            $\beta$ & The distance threshold of neighboring vertices \\ \hline
            $\theta$ & The number threshold of vertices in the subgraph \\ \hline
            $p_i$ & The pixel coordinates of vertex $v_i$ in the image \\ \hline
            $\tilde{p_i}$ & The centroid of subgraph $G_i$ \\ \hline
            $\mathbf{S}$ & The score matrix of vertices \\ \hline
            $\mathbf{\tilde{S}}$ & The matrix $\mathbf{S}$ with the dustbin \\ \hline
            $\mathbf{Q}$ & The assignment matrix of vertices \\ \hline
            $\mathbf{\tilde{Q}}$ & The matrix Q with the dustbin \\ \hline
            $\mathbf{f_i}$ & The feature descriptor of vertex $v_i$  \\ \hline
            $\mathbf{\tilde{f}_i}$ & The feature incorporating $\mathbf{f_i}$ and $MLP(p_i)$  \\ \hline
            $\mathbf{\tilde{f}_i^A}$ & The $\mathbf{\tilde{f}_i}$ in image $A$ \\ \hline
            $h_i^k$ & The embedding of vertex $v_i$ at the $k$-th iteration in GNN \\ \hline
            $\sigma$ & The activation function in GraphSAGE \\ \hline
            $\mathbf{W}$ & The learned weight matrix in GraphSAGE \\
             \hline
        \end{tabular}%
    }
    \vspace{-1.8em}
\end{table}

\subsubsection{\textbf{Keypoint Detection}}
In image matching, vertices typically represent keypoints in an image, while edges between vertices represent spatial or visual relationships between keypoints. Note that we use the term ``vertex'' consistently throughout this paper. Let $V$ be the vertex set and $E$ be the edge set, the graph $G$ can be represented as $G = (V, E)$.

In the proposed system, any keypoint detection method, such as SIFT, ORB, or SuperPoint, can be used to generate keypoints, which correspond to vertices in the graph. However, we recommend using a detector that provides a response for each keypoint, such as the SIFT detector used in this work. To generate robust local descriptors, we adopt CAR-HyNet~\citep{song2023carhynet} to extract enhanced patch-level representations.

\subsubsection{\textbf{Adaptive Coarse Graph Construction}}
We first extract feature descriptors of vertices and calculate the cosine similarity between them. For SIFT descriptors $\mathbf{d}_i$ and $\mathbf{d}_j$ of two vertices $v_i$ and $v_j$, the cosine similarity is given by,
\begin{equation}
	\label{eq:cosine_similarity}
	c_{ij} = \frac{\mathbf{d}_i \cdot \mathbf{d}_j}{\|\mathbf{d}_i\|_2 \|\mathbf{d}_j\|_2},
\end{equation}
where $\mathbf{d}_i \cdot \mathbf{d}_j$ denotes the dot product, $\|\mathbf{d}_i\|_2$ and $\|\mathbf{d}_j\|_2$ denote the Euclidean norms of $\mathbf{d}_i$ and $\mathbf{d}_j$, respectively.

Let \( \mathbf{C} = \{ c_{ij} \mid 1 \le i \leq j \le n \} \) represent the upper triangular matrix of cosine similarities for all vertex pairs, excluding the diagonal, where \( n \) is the total number of vertices. To adapt the similarity threshold to graph variations, we flatten and sort \( \mathbf{C} \) to obtain \(\tilde{C } = \{ c_{(1)}, c_{(2)}, \ldots, c_{(k)} \}\), where \(k = \frac{n(n-1)}{2}\) is the number of elements in the upper triangular matrix \( \mathbf{C} \). Given the percentile threshold $\alpha$, we calculate the corresponding cosine similarity threshold $\gamma$ for each graph as,
\begin{equation}
	\label{eq:cosine_similarity}
    \gamma = \big(1 - \{r\}\big) \times c_{\lfloor r \rfloor} + \{r\} \times c_{\lfloor r \rfloor + 1},
\end{equation}
where \(r = \frac{\alpha}{100} \times (k - 1)\) represents the index corresponding to the percentile threshold $\alpha$ in $\tilde{C}$, $\lfloor r \rfloor$ denotes the integer part of $r$, \{r\} denotes the fractional part of $r$, and $c_{(\lfloor r \rfloor)}$ denotes the $\lfloor r \rfloor$-th element in $\tilde{C}$. The preset $\alpha$ is used to localize the threshold in the similarity distribution, while $\gamma$ is dynamically calculated from each graph, ensuring adaptability. 

KDTree~\citep{bentley1975multidimensional} is a data structure for organizing vertices in the K-dimensional space that can efficiently find the neighboring vertices of each vertex in the graph. To construct the coarse graph, we first find all vertices within a specific range for each vertex using KDTree. For vertex $v_i$, we find its neighboring vertices set $N(i)$ by,
\begin{equation}
	\label{eq:neighboring}
	N(i) = \{v_j \in V | \|v_i - v_j\| < \beta, v_i \neq v_j\},
\end{equation}
where \(\|v_i - v_j\|\) denotes the Euclidean distance between vertices $v_i$ and $v_j$, and $\beta$ is the spatial proximity threshold.

With the above steps, we start to construct a graph that takes into account both spatial proximity and feature similarity. For each vertex $v_i$ and its neighbor $v_j$, an edge is added between them if the cosine similarity $c_{ij}$ between them is greater than or equal to the threshold $\gamma$ given by \myeq{eq:cosine_similarity}. In other words, the edge $e_{ij}$ is added for all vertex pairs $(v_i, v_j)$ that satisfy the following condition,
\begin{equation}
	\label{eq:edge-addition-rule}
	\text{if } c_{ij} \geq \gamma \text{ and } \|v_i - v_j\|_2 \leq \beta, \text{ add edge } e_{ij}.
\end{equation}

\subsubsection{\textbf{Connection of Isolated Vertices}}
Obviously, distance-based graph construction methods result in numerous isolated vertices, that is, vertices without neighbors. To minimize the impact of the preset distance radius $\beta$ and considering that simply ignoring isolated vertices would directly lead to the loss of useful information in the graph representation. Therefore, we aim to include as many isolated vertices as possible.

To achieve this goal, based on $\gamma$ and $\beta$ in \myeq{eq:cosine_similarity} and \myeq{eq:neighboring}, we use the following method to create edges to connect isolated vertices to their nearest neighboring vertices, thus integrating them into existing subgraphs or forming new subgraphs. Specifically, for each isolated vertex $v_i$, its nearest neighbors are queried using KDTree. Let $p_i$ be the pixel coordinate of the vertex $v_i$ in the image, then for an isolated vertex $v_i$, we look for the nearest vertex $v_j$ that satisfies \myeq{eq:connect_isolated_nodes}, and an edge is added between $v_i$ and $v_j$,
\begin{equation}
	\label{eq:connect_isolated_nodes}
	p_j = \underset{v_t \in V \setminus \{v_i\}}{\mathrm{argmin}}\, \lVert p_i - p_t \rVert_2,
\end{equation}
where $\lVert \cdot \rVert_2$ denotes the Euclidean distance, and $p_j$ is the the pixel coordinate of vertex $v_j$.

\subsubsection{Removal of Small Subgraphs}

To improve efficiency and representation quality, we remove small subgraphs that are structurally isolated from the main graph. Although previous step reduces the number of isolated vertices, it may still result in several disconnected subgraphs that often comprise only a few vertices due to spatial dispersion. These small components are usually internally connected but remain disjoint from the global structure. Such subgraphs often correspond to peripheral or noisy regions and contribute little to the overall matching task. Attempting to reconnect all subgraphs increases algorithmic complexity without guaranteed performance gain, while leaving them intact incurs unnecessary computation. 

We first represent graph $G$ as a union of all its disjoint subgraphs, i.e., subgraphs that are not connected to each other,
\begin{align}	
\label{eq:disjoint_subgraphs}
    G=\bigcup_{\forall G_i \subseteq G} G_i,
\end{align}
where for any $G_i$ and $G_j$ in \myeq{eq:disjoint_subgraphs}, $G_i\cap G_j=\emptyset$. 

Next, we use the graph order $|V|$ as a measurement for the removal of small subgraphs. Specifically, for a disjoint subgraph $G_i=(V_i, E_i)$ in \myeq{eq:disjoint_subgraphs}, we remove $G_i$ from $G$ if its order is smaller than a given threshold $\theta$ as follows,
\begin{align}	
\label{eq:remove_small_subgraphs}
     G = G \setminus G_i, \text{if } G\neq G_i\text{ and } |V_i| < \theta, 
\end{align}
where $|V_i|$ denotes the order of $G_i$, $\theta$ is the minimum order for any subgraph $G_i$ required to be kept in $G$. We can further flexibly control the density of the graph by setting $\theta$.

Eventually, we have $G$ either as a single connected graph or a set of disjoint subgraphs with orders larger than $\theta$, i.e.,
\begin{align}	
\label{eq:coarse_graph}
    G=\bigcup_{\forall G_i\subseteq  G} G_i\text{, where } 
     |V_i|\geq \theta.
\end{align}

\subsubsection{Fine Graph Construction}
Finally, we ensure that the final graph $G=(V, E)$ is connected, i.e., $\forall v_i\in V$, there $\exists v_j \in V\setminus \{v_i\}$ where $e_{ij}\in E$. To achieve this, we create edges to connect any disjoint subgraphs from \myeq{eq:coarse_graph} as follows. 

First, we calculate the centroid for each subgraph using \myeq{eq:centroid} to represent the current subgraph,
\begin{equation}
	\label{eq:centroid}
    \tilde{p}_i = \frac{1}{|V_i|} \sum_{v_j \in V_i} p_j,
\end{equation}
where $\tilde{p}_i$ denotes the centroid coordinates of the $i$-th subgraph, and $p_j$ denotes the pixel coordinates of vertex $v_j$ in $V_i$.

Then, based on the centroid coordinates, we use KDTree to find the nearest neighbors of each subgraph. After finding the nearest centroid pairs, we search for the nearest vertices within the corresponding subgraphs. Specifically, for the centroid pair $(\tilde{p}_i, \tilde{p}_j)$, we find the nearest vertex pair $(v_i, v_j)$ in the corresponding subgraphs by \myeq{eq:connect_isolated_nodes} and add an edge \( E = E \cup \{e_{ij}\} \).

This process is repeated until the graph $G$ is a connected graph, i.e., $\forall v_i\in V$, $\exists v_j\in V\setminus \{v_i\}$, $e_{ij} \in E$.

\subsubsection{Complexity Analysis.}
\label{sec:complexity}
We analyze the time complexity of each step in our AGC algorithm. Let $n$ denote the number of detected keypoints, $f$ denote the descriptor dimension, $d$ denote the patch size, $k$ denote the average number of spatial neighbors, and $t$ denote the number of connected components during refinement, and $W \times H$ denote the input image size. In typical settings, we assume $k, d, f, t \ll n$.

\begin{itemize}[leftmargin=1.2em]
    \item \textbf{Keypoint Detection.} 
    We use SIFT to detect $n$ keypoints from an image of size $W \times H$, involving Gaussian pyramid construction and orientation detection, with cost $\mathcal{O}(WH +n)$. For each keypoint, a $d \times d$ patch is encoded by CAR-HyNet, with total cost $\mathcal{O}(n d^2)$. Thus, the overall complexity of this stage is $\mathcal{O}(WH + n d^2)$.

    \item \textbf{Adaptive Coarse Graph Construction.} 
    KDTree construction takes $\mathcal{O}(n \log n)$, neighbor queries cost $\mathcal{O}(nk+n \log n)$, and cosine similarity for $n \times k$ pairs requires $\mathcal{O}(n k f)$. Threshold selection using partial sorting or selection adds $\mathcal{O}(n k)$. As $k \ll n$, the total complexity of this step is $\mathcal{O}(n \log n + nkf)$.
        
    \item \textbf{Connection of Isolated Vertices.} 
    For each isolated vertex, we query the nearest neighbor using KDTree in $\mathcal{O}(\log n)$ time. In the worst case, where all $n$ vertices are isolated, yielding total complexity $\mathcal{O}(n \log n)$.

    \item \textbf{Removal of Small Subgraphs.} 
    We identify connected components using BFS, with total cost $\mathcal{O}(n + |E|)$ for a graph with $n$ vertices and $|E|$ edges. Filtering and removing subgraphs smaller than $\theta$ is a linear pass over all components. Thus, the total complexity of this step is $\mathcal{O}(n + nk)$.

    \item \textbf{Fine Graph Construction.} 
    To connect $t$ disjoint subgraphs, we first compute their centroids in $\mathcal{O}(n)$ time. Finding the closest centroid pairs via KDTree takes $\mathcal{O}(t \log t)$, and searching for the closest vertex pair across two subgraphs takes up to $\mathcal{O}(n \log n)$ in the worst case. Repeating this for up to $t - 1$ rounds gives a total cost of $\mathcal{O}(nk + tn \log n)$, which is $\mathcal{O}(n^2 \log n)$ in the worst case when $t=n$.
\end{itemize}

In sparse graphs, the overall time complexity of AGC is $\mathcal{O}(WH + nd^2 + nkf + tn \log n)$.

\begin{figure*}[tbp]
    \centering
    \setlength{\abovecaptionskip}{0pt}
    \setlength{\belowcaptionskip}{0pt}
    \includegraphics[width=0.9\linewidth]{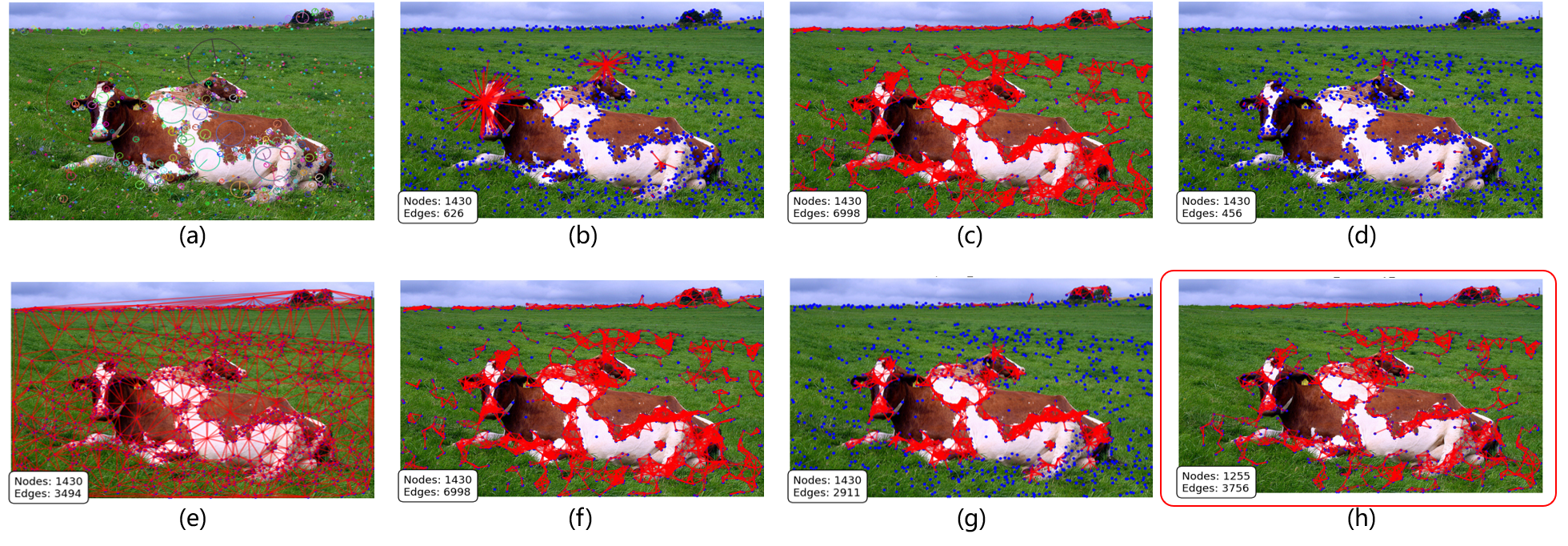}
    \caption{\textbf{Comparison of different graph construction methods.} (a) SIFT keypoints, (b) connect vertices within a circle defined by the size of SIFT keypoints, (c) connect vertices within a fixed distance threshold, (d) the smaller of the size of SIFT keypoints or a specified distance threshold as the distance threshold, (e) use Delaunay triangulation, (f) connect vertices with stronger intensity of keypoints within a fixed distance threshold, (g) connect vertices with intensity of keypoints higher than the local average intensity within a fixed distance threshold, and (h) connect vertices based on the distance and similarity-aware adaptive graph construction method we employ.}
    \label{fig:cgm}
    \vspace{-1em}
\end{figure*}

\subsubsection{Comparison of Different Graph Construction Methods}
We compare several methods of graph construction, the results are shown in Fig.~\ref{fig:cgm}, where Fig.~\ref{fig:cgm}(a) shows keypoints detected by SIFT, and Fig.~\ref{fig:cgm}(b)-(h) show the graph construction based on various vertex connection rules. It is observed that some methods result in many isolated vertices, as shown in Fig.~\ref{fig:cgm}(b) and Fig.~\ref{fig:cgm}(d). This affects the performance of graph algorithms, as the graph is very limited in the information it can provide about its neighbors. Other methods result in many isolated subgraphs, as shown in Fig.~\ref{fig:cgm}(c), Fig.~\ref{fig:cgm}(f), and Fig.~\ref{fig:cgm}(g). Smaller subgraphs contain less information and affect the efficiency of the algorithm. Meanwhile, Fig.~\ref{fig:cgm}(e) generates a fully connected graph that does not effectively reflect the main structure. In contrast, our method, as shown in Fig.~\ref{fig:cgm}(h), maintains the main structure of the image well while reducing the number of irrelevant vertices and edges. This balanced connectivity is conducive to the accuracy of subsequent image analysis tasks, as it preserves essential features while minimizing noise and computational complexity.

\subsection{GNN Based Graph Matching}
\begin{figure*}[!htb]
    \vspace{-1em}
    \centering
    \setlength{\abovecaptionskip}{0pt}
    \setlength{\belowcaptionskip}{0pt}
    \includegraphics[width=0.95\linewidth]{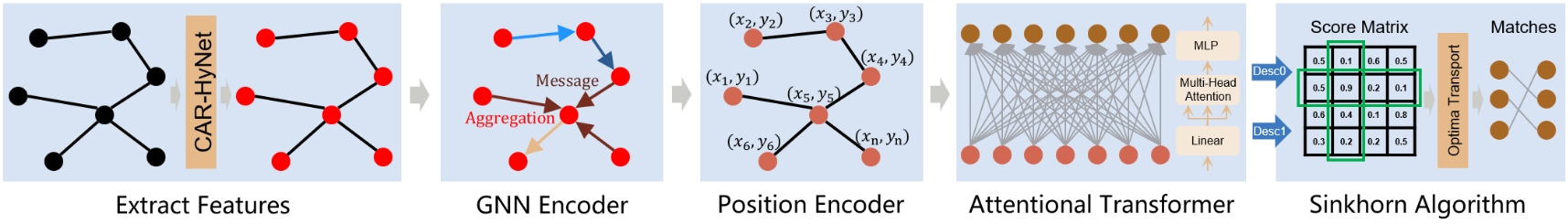}
    \caption{\textbf{Overall flow of graph matching based on GNN.} For the graphs obtained from \textit{Adaptive Graph Construction}, we extract vertex-centered patches that are fed into CAR-HyNet~\citep{song2023carhynet} to extract feature descriptors. The GNN encodes the local spatial information of vertices and integrates the positional data, while the Transformer encodes the global information. By computing the inner product of the feature descriptors of two graphs, we generate and optimize the score matrix with Sinkhorn's algorithm to produce a matching result.}
    \label{fig:graph_matching_overall_flow}
    \vspace{-1em}
\end{figure*}

In this section, we describe the graph matching process and the effective integration of GNN and Transformer in detail. Fig.~\ref{fig:graph_matching_overall_flow} shows the overall flow of GNN-based graph matching. 

\subsubsection{Feature Extraction for Graph Vertices}
We mimic the SIFT method to reconstruct the Gaussian scale pyramid. Then, based on the keypoint information in the scale space corresponding to each keypoint, we crop 64$\times$64 images centered on keypoints and resize them to 32$\times$32 to form patches. However, unlike SIFT, we handle color images here. Furthermore, in order to obtain rotational invariance, we rotate the patches according to the orientations of the SIFT keypoints.

\subsubsection{GNN Local Spatial Encoder}
For GNNs, the graph is updated by iteratively updating the representation of each vertex. The update process produces vertex embeddings that reflect the local graph topology and vertex characteristics based on the features of their neighbors and the vertices themselves. The update rule can commonly be formulated as,
\begin{equation}
    \label{eq:gnn_formula}
	\begin{aligned}
		\resizebox{0.7\hsize}{!}{
			   $h_i^{(k+1)} = \text{UPDATE}\left(h_i^{(k)}, \text{AGGREGATE}\left(\{h_j^{(k)} : j \in N(i)\}\right)\right)$,
		}
	 \end{aligned}
\end{equation}
where $h_i^{(k)}$ is the embedding of vertex $v_i$ at the $k$-th iteration and $N(i)$ denotes the neighbors of vertex $v_i$. The \textit{AGGREGATE} function combines the embeddings of these neighbors, and the \textit{UPDATE} function generates new embeddings by combining the previous embeddings of the vertex with the aggregated neighborhood information.

GNNs encompass various models, such as Graph Attention Networks (GAT)~\citep{velickovic2018graph}, Graph Convolutional Network (GCN)~\citep{gcn2017}, and GraphSAGE~\citep{graphsage2017}. In this work, we focus on GraphSAGE, which generates embedded representations of vertices by sampling and aggregating feature information from neighboring vertices. This approach makes GraphSAGE suitable for processing large-scale graph data. Moreover, the scope of information propagation can be effectively controlled by adjusting the network depth, i.e., the number of hops. With each additional layer, the model can aggregate information from further neighbors. GraphSAGE is defined as,
\begin{equation}
    \label{eq:graphsage_formula}
	\begin{aligned}
		\resizebox{0.7\hsize}{!}{$
                 h_i^{(k+1)} = \sigma\left(\mathbf{W} \cdot \text{MEAN}\left(\{h_i^{(k)}\} \cup \{h_j^{(k)} : \forall j \in N(i)\}\right)\right),
            $}
	 \end{aligned}
\end{equation}
where $\sigma$ denotes the activation function, $\mathbf{W}$ is the learned weight matrix, $\text{MEAN}(\cdot)$ represents the mean-value aggregation function, which averages the embedding vector of vertex $v_i$ with embedding vectors of all its neighbors.

After GNN encoding, each vertex in the graph acquires a comprehensive feature representation that includes its neighborhood information. However, encoding too many neighborhood hops can lead to over-smoothing. This occurs when repeated aggregation causes the information from different vertices to mix excessively, making their feature representations more similar. This reduces the distinctiveness of vertex features, thus reducing the effectiveness of embeddings and negatively impacting GNN performance in downstream tasks. Our experiments indicate that optimal performance is achieved by selecting neighbors with three hops.

\subsubsection{Vertex Position Encoder}
The GNN encoder enables each vertex to incorporate the information of its neighbors, while the positional encoding integrates the relative positional relationships of a vertex with its neighbors. We use an MLP to encode each vertex position and sum the encoding result with the vertex feature to obtain a new feature representation,
\begin{equation}
	\label{eq:position_encoder}
    \mathbf{\tilde{f}_i} = \mathbf{f_i} + \text{MLP}(p_i),
\end{equation}
where $\mathbf{\tilde{f}_i}$ is the new feature representation of vertex $v_i$ incorporating position information, $\mathbf{f_i}$ is the initial feature representation of vertex $v_i$ obtained by SIFT, and $p_i$ is the position information of vertex $v_i$. MLP($\cdot$) is a multilayer perceptron that takes the position information as an input and outputs a feature vector that will be added to the initial feature vector to get the final vertex feature representation.

\subsubsection{Attentional Transformer Encoder}
The Transformer architecture captures global dependencies, enabling the model to perform global feature interactions and information propagation across the entire graph. We adopt the design of self-attention and cross-attention for the Transformer as described in~\citep{sarlin2020superglue}. The self-attention and cross-attention layers are alternated, mimicking the process that humans would use against each other when making a match.

\subsubsection{Graph Matching Module}
Given two graphs $G^A = (V^A, E^A)$ and $G^B = (V^B, E^B)$, 
the goal of Sinkhorn-based graph matching is to compute a soft assignment matrix $\mathbf{Q} \in \mathbb{R}^{m \times n}$ to establish vertex correspondences. Following standard practice~\citep{cuturi2013sinkhorn, sarlin2020superglue}, we first compute a score matrix $\mathbf{S}$ based on descriptor similarity as $s_{ij} = \langle \tilde{\mathbf{f}}_i^A, \tilde{\mathbf{f}}_j^B \rangle$, and augment it with a ``dustbin'' row and column to handle unmatched nodes. We then apply the Sinkhorn algorithm to obtain a doubly-stochastic matrix $\tilde{\mathbf{Q}} = \mathrm{Sinkhorn}(\tilde{\mathbf{S}})$. Compared to the exponential complexity of the brute-force method, this approach provides a differentiable and efficient solution for soft graph matching.

\subsubsection{Optimisation of Matching Results}
The homography matrix~\citep{fischler1981random} is often used for refining feature matching results. The RANSAC algorithm and its variants are commonly used to estimate the homography matrix by iteratively selecting and verifying keypoint correspondences, effectively filtering out outliers. We apply it to filter the matching vertices.


\subsection{Data Parallel Multi-GPU Training}
In deep learning, parallel training can be primarily divided into two types: \textit{Model Parallelism} and \textit{Data Parallelism}. Given that our model has 12 million parameters with a memory footprint of 1 GB, it can be fully loaded and trained on a single device. However, large datasets can significantly reduce training efficiency, such as the 118k images as the training set of COCO2017~\citep{COCO2017} dataset. Therefore, we consider employing data parallelism to partition the dataset into multiple small batches and process these batches concurrently on multiple GPUs to enhance the training efficiency and reduce training time. Note that distributed training across multiple machines and GPUs can be readily achieved with minor adjustments.

\section{Experiments}
\label{sec:experiments}
In this section, we describe the lab setup, the datasets, and the evaluation criteria used in the experiments. Then, we conduct experiments on public datasets and real-world images and compare the proposed image matching system with existing methods.

\subsection{Lab Setup and Datasets}
\begin{figure}[htb]
    \vspace{-0.5em}
    \centering
    \setlength{\abovecaptionskip}{0pt}
    \setlength{\belowcaptionskip}{0pt}
    \includegraphics[width=\linewidth]{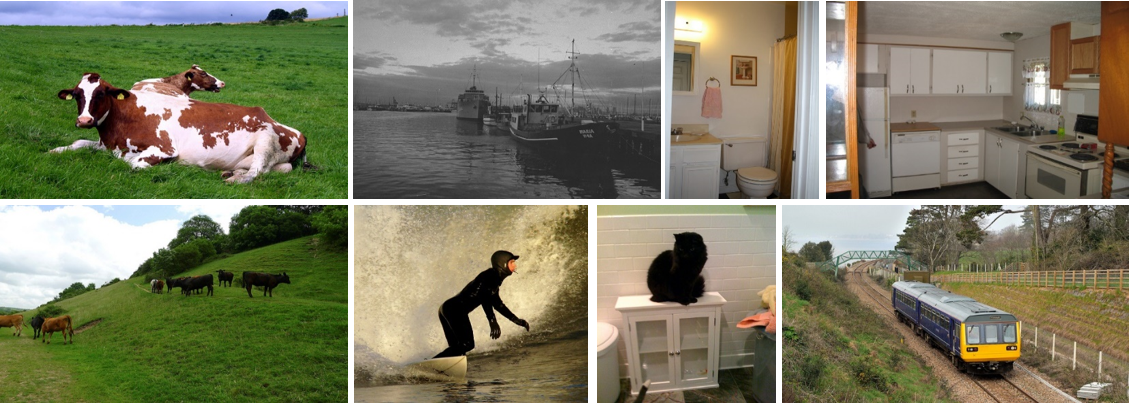}
    \caption{\textbf{Samples from COCO2017 dataset.}}
    \label{fig:samples_from_COCO2017_dataset}
    \vspace{-0.5em}
\end{figure}

We use the large image dataset COCO2017~\citep{COCO2017} to train our model on the \textit{Keypoint} task. This dataset contains 118,287 images in the \textit{training set}, 5,000 images in the \textit{validation set}, and 40,670 images in the \textit{test set}. Some image samples are shown in Fig. \ref{fig:samples_from_COCO2017_dataset}. We evaluate on a selection of images from the test set in the COCO2017 dataset, the \textit{indoor test set} and \textit{outdoor test set} in the RGB-D dataset ~\citep{rgb_d} and the Oxford-Affine dataset~\citep{mikolajczyk2005comparison}. We also conduct experiments on real-world images. We choose a 3-layer GraphSAGE as the GNN model and train it for two epochs. During training, we obtain pairs of images to be matched and their corresponding true matching keypoints by generating random homographies that include transformations such as scaling, rotation, perspective, cropping, and translation. As a common practice, we scale images to 800$\times$600. Finally, we also explore the impact of different numbers and types of GPUs on training acceleration.

We compare our system with several classical methods, and use \textit{Area under Curve} (AUC) of the cumulative pose estimation error and \textit{Match Number} (MN) of image pairs as evaluation metrics. The combinations of different methods are presented in Table \ref{table:use_of_different_combinations_of_algorithms}. Moreover, during inference, we limit the maximum number of keypoints that can be detected by each algorithm to 10,000. For keypoints that exceed this limit, we sort them by their responses or scores and select the top 10,000 keypoints. For simplicity, we assign a label to each method and use these labels consistently throughout the paper. Note that, since SuperGlue directly applies the Transformer to keypoints, it can be considered to construct a complete graph. Also note that for traditional methods, we use the \textit{Nearest Neighbor Distance Ratio} (NNDR)~\citep{lowe2004distinctive} as the matcher. For deep learning models such as OmniGlue\footnote{\url{https://github.com/google-research/omniglue/}}, DeDoDe\footnote{\url{https://github.com/Parskatt/DeDoDe}}, SuperPoint\footnote{\url{https://github.com/magicleap/SuperPointPretrainedNetwork}} and SuperGlue\footnote{\url{https://github.com/magicleap/SuperGluePretrainedNetwork}}, we use their officially provided pre-trained weights and recommended parameters. Note that for the \textit{SCN}, we do not use the feature fusion method in ~\citep{song2023carhynet}.

\begin{table}[!htb]
    \vspace{-1em}
    \setlength{\abovecaptionskip}{0pt}
    \setlength{\belowcaptionskip}{0pt}
    \centering
    \caption{\textbf{Combination of different methods.} For deep learning models, we use their officially provided pre-trained models.}
    \resizebox{\columnwidth}{!}{%
        \renewcommand{\arraystretch}{1.2}
        \begin{threeparttable}
            \begin{tabular}{lcccc}
                \Xhline{1.5pt}
                \noalign{\vskip 2pt}
                \textbf{Label} & \textbf{Graph} & \textbf{Detector} & \textbf{Descriptor} & \textbf{Matcher} \\ 
                \noalign{\vskip 2pt}
                \Xhline{1pt}
                \noalign{\vskip 3pt}
                SGO~\citep{sarlin2020superglue} & Complete & \makecell[c]{SuperPoint\\\citep{detone2018superpoint} \\ (v1 model)} & \makecell[c]{SuperPoint \\ (v1 model)} & \makecell[c]{SuperGlue \\ (outdoor model)} \\ 
                \noalign{\vskip 3pt}
                SGI~\citep{sarlin2020superglue} & Complete & \makecell[c]{SuperPoint \\ (v1 model)} & \makecell[c]{SuperPoint \\ (v1 model)} & \makecell[c]{SuperGlue \\ (indoor model)} \\ 
                \noalign{\vskip 3pt}
                OmniGlue~\citep{jiang2024Omniglue} & Complete & OmniGlue  &  OmniGlue & OmniGlue \\
                DeDoDe~\citep{edstedt2024dedode} & None & \makecell[c]{DeDoDe \\ (detector\_L)} &  \makecell[c]{DeDoDe \\ (descriptor\_B)} & \makecell[c]{DeDoDe \\ (DualSoftMax)} \\ 
                \noalign{\vskip 3pt}
                SSN~\citep{lowe1999object}    & None     & SIFT & SIFT & NNDR~\citep{lowe2004distinctive} \\ 
                SCN~\citep{song2023carhynet}    & None     & SIFT & CAR-HyNet & NNDR \\ 
                D-GIMS$^*$ & Delaunay & SIFT & CAR-HyNet & GNNMatcher \\
                \textbf{GIMS$^*$} & \textbf{AGC} & \textbf{SIFT} & \textbf{CAR-HyNet} & \textbf{GNNMatcher} \\ 
                \Xhline{1.5pt}
            \end{tabular}
            \begin{tablenotes}
    		  \item $^*$This Work.
         \end{tablenotes}
        \end{threeparttable}
    }
    \label{table:use_of_different_combinations_of_algorithms}
\end{table}

We use the following software and hardware resources for the above evaluation. The graph-based image matching system runs on a Supermicro server, equipped with a dual-core Intel(R) Xeon(R) Gold 6230 CPU, two RTX 3090 GPUs, and two Tesla A40 GPUs, and 754GB of RAM. The software environment includes Ubuntu 20.04, Python 3.8, PyTorch 2.0.1, and DGL 2.0.0.

\subsection{Parameter Analysis}
As described in Section~\ref{sec:method}, the GIMS system introduces three key parameters in graph construction: the neighbor radius $\beta$, similarity threshold $\alpha$, and minimum subgraph size $\theta$. These correspond to spatial constraints, edge filtering strength, and noise suppression, jointly affecting graph connectivity, structural integrity, and stability, thereby balancing matching performance and computational cost.

\begin{figure*}[!htb]
    \centering
    \setlength{\abovecaptionskip}{0pt}
    \setlength{\belowcaptionskip}{0pt}
    \includegraphics[width=\linewidth]{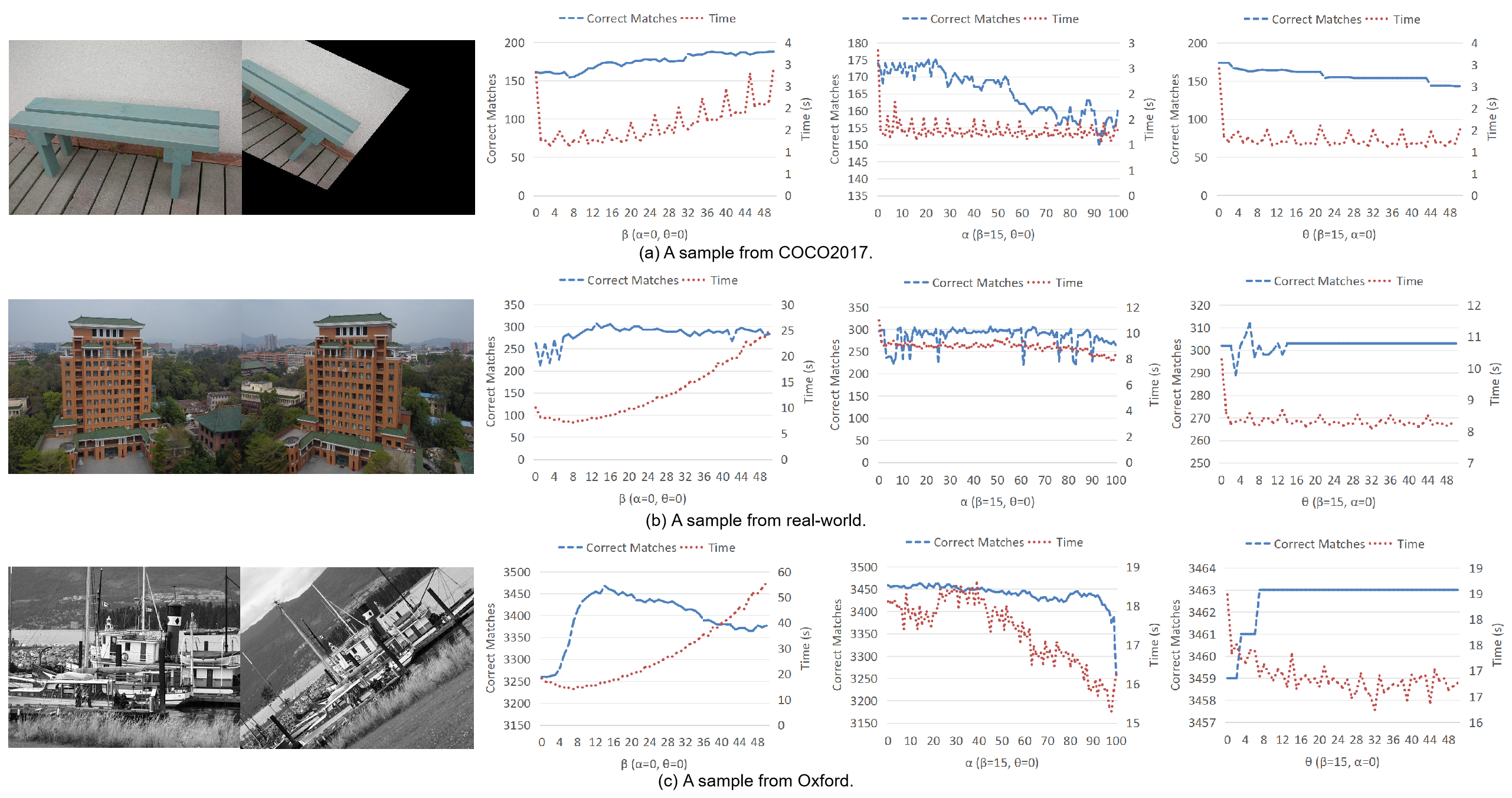}
    \caption{\textbf{Effect of parameters $\beta$, $\alpha$ and $\theta$.} }
    \label{fig:parameter1}
    \vspace{-1em}
\end{figure*}

Fig.~\ref{fig:parameter1} shows how these parameters influence the number of correct matches and runtime. Increasing $\beta$ expands connectivity and yields more potential matches but introduces redundant edges, raising the risk of false matches and computational overhead. Reducing $\beta$ simplifies the graph but may lead to fragmentation in sparse areas. Lower $\alpha$ preserves more edges and ensures global connectivity, while higher $\alpha$ favors reliable connections and improves efficiency. Increasing $\theta$ removes weak subgraphs and speeds up processing, though overly large values may prune fine-grained structures, causing fragmentation and extra reconstruction cost.

In practice, the optimal parameter choices can vary with scene characteristics. For example, in texture-sparse and structurally simple images as shown in Fig.\ref{fig:parameter1}(a), decreasing $\alpha$ and $\theta$ while increasing $\beta$ helps preserve essential connections and avoids over-filtering. In contrast, for densely textured scenes as shown in Fig.\ref{fig:parameter1}(b) and Fig.\ref{fig:parameter1}(c), increasing $\alpha$ and $\theta$ effectively reduces runtime with little or even positive impact on accuracy. 

\begin{figure*}[!htb]
    \centering
    \setlength{\abovecaptionskip}{0pt}
    \setlength{\belowcaptionskip}{0pt}
    \includegraphics[width=\linewidth]{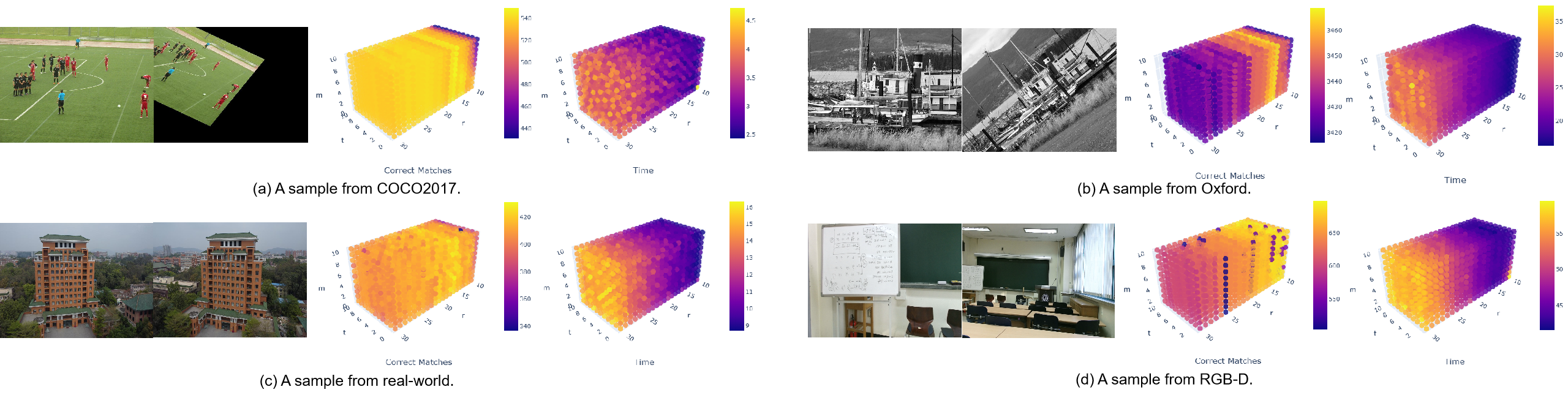}
    \caption{\textbf{Grid search for parameters $\beta$, $\alpha$ and $\theta$.} }
    \label{fig:parameter2}
    \vspace{-1em}
\end{figure*}

To enhance the generality and robustness of the system, we perform a grid search over the parameter ranges $\beta \in [10, 30]$, $\alpha \in [0, 10]$, and $\theta \in [0, 10]$, and systematically evaluate the matching performance across multiple image pairs, as shown in Fig.~\ref{fig:parameter2}. The final configuration $(15, 2, 7)$ is selected as a balanced setting, demonstrating good performance and generalization in most scenarios in the experiments.

\subsection{Evaluation of Pose Estimation Accuracy with AUC}
As a common practice, we use the AUC to evaluate the accuracy of pose estimation and quantify the performance of the matching method at error thresholds of 5, 10, and 25 pixels. The AUC provides a measure of the overall performance of the model, with higher AUC values indicating greater accuracy and robustness. Based on the matching results, we use the RANSAC algorithm to calculate the homography matrix. The AUC at different thresholds can be obtained by calculating the average re-projection error at four corners of the image based on the true homography matrix.

\begin{table*}[!htbp]
    \setlength{\abovecaptionskip}{0pt}
    \setlength{\belowcaptionskip}{0pt}
    \centering
    \caption{\textbf{Pose Estimation with AUC (@5/@10/@25) and Average Match Number (AMN)}. GIMS outperformes others in all scenarios.}
    \label{table:pose_estimation_auc}
    \resizebox{\linewidth}{!}{%
        \setlength\tabcolsep{10pt}
        \renewcommand{\arraystretch}{1.3}
        \begin{threeparttable}
            \begin{tabular}{lcccccccccccc}
                \Xhline{1.5pt}
                \noalign{\vskip 2pt}
                \multirow{2}{*}{\textbf{Label}} &
                  \multicolumn{4}{c}{\textbf{Test (COCO2017)}} &
                  \multicolumn{4}{c}{\textbf{Indoor (RGB-D)}} &
                  \multicolumn{4}{c}{\textbf{Outdoor (RGB-D)}} \\
                  \cmidrule(lr){2-5}\cmidrule(lr){6-9}\cmidrule(lr){10-13}
                  & \textbf{@5} & \textbf{@10} & \textbf{@25} & \textbf{AMN} & \textbf{@5} & \textbf{@10} & \textbf{@25} & \textbf{AMN} & \textbf{@5} & \textbf{@10} & \textbf{@25} & \textbf{AMN} \\
                \Xhline{1pt}
                SGO~\citep{sarlin2020superglue} & 33.75 & 50.52 & 71.16 & 208 & 30.47 & 44.78  & 61.10 & 211 & 30.85 & 42.14 & 55.79 & 344 \\
                SGI~\citep{sarlin2020superglue} & 31.60 & 47.97 & 68.27 & 187 & 26.68 & 40.72 & 56.58 & 176 & 26.55 & 38.28 & 51.96 & 264 \\
                OmniGlue~\citep{jiang2024Omniglue} & 35.12 & 51.96 & 71.01 & 349 & 33.62 & 45.88 & 61.08 & 311 & 29.77 & 40.69 & 52.57 & 419 \\
                DeDoDe~\citep{edstedt2024dedode}  & 56.15 & 68.99 & 80.16 & 957 & 49.80 & 59.50 & 69.46 & 601 & 48.36 & 56.64 & 65.21 & 616 \\
                SSN~\citep{lowe1999object}    & 65.18 & 77.23 & 87.20 & 418 & 47.12 & 56.96  & 67.38 & 163 & 42.90 & 50.94 & 59.88 & 229 \\
                SCN~\citep{song2023carhynet}  & 66.45 & 77.93 & 87.92 & 312 & 50.22 & 60.14  & 71.37 & 130 & 42.95 & 51.80 & 61.61 & 201 \\
                D-GIMS$^*$   & 77.71 & 86.43 & 92.37 & 1483 & 62.33 & 69.48 & 75.92 & 1103 & 50.84 & 57.08 & 63.43 & 966 \\
                \textbf{GIMS$^*$} &
                  \textbf{78.63}  &
                  \textbf{87.74}  &
                  \textbf{93.92}  &
                  \textbf{1723}   &
                  \textbf{64.70}  &
                  \textbf{72.77}  &
                  \textbf{79.58}  &
                  \textbf{1216}   &
                  \textbf{53.85}  &
                  \textbf{60.05}  &
                  \textbf{65.83}  &
                  \textbf{1043}   \\
                \Xhline{1.5pt}
            \end{tabular}%
            \begin{tablenotes}
    		  \item $^*$This Work.
            \end{tablenotes}
        \end{threeparttable}
    }
    \vspace{-1em}
\end{table*}

We randomly select 1,000 images from the COCO2017 test set. In addition, we select all 503 images in the indoor test set and all 500 images in the outdoor test set of the RGB-D dataset. Note that the actual number of valid images may vary slightly because some algorithms fail to detect matching keypoints on certain images, resulting in the inability to calculate the AUC. As can be seen in Table~\ref{table:pose_estimation_auc}, GIMS achieves the highest AUC among all comparison methods. Specifically, the COCO2017 dataset contains various categories of images, including daily scenes, objects, and faces. GIMS achieves a significantly higher AUC on the COCO2017 test set. Indoor images usually face challenges such as lack of texture and scene complexity. GIMS still achieves the highest AUC performance under all three thresholds. Additionally, outdoor images often face multiple challenges, such as changes in lighting. According to the results, SCN achieves a higher AUC than SGI, SGO, OmniGlue, SSN, and DeDoDe under three thresholds, while D-GIMS and GIMS further improve their AUC. This is partly because OmniGlue and DeDoDe exhibit weaker robustness in certain challenging scenes such as strong perspective transformations and low-texture regions. More specifically, D-GIMS and GIMS use SIFT to detect keypoints and CAR-HyNet to generate keypoint features. Therefore, the excellent performance of D-GIMS and GIMS can be attributed in part to the strengths of these two algorithms. This is evident from SCN achieving a better AUC than all other methods except D-GIMS and GIMS. However, D-GIMS and GIMS further enhance their strengths by leveraging the characteristics of GNN and Transformer. 

On the other hand, D-GIMS uses Delaunay triangulation to construct the graph, while GIMS uses AGC. The graph constructed by Delaunay triangulation is too dense and poorly reflects the main structure. Thus, although D-GIMS, with the help of SIFT, CAR-HyNet, and GNNMatcher, achieves a higher AUC compared to other methods, it still falls short of the proposed GIMS, demonstrating the effectiveness of the AGC method. In general, the experimental results demonstrate that GIMS is robust to challenges such as illumination changes, occlusions, and viewpoint variations. GIMS significantly improves the accuracy of pose estimation and effectively adapts to different scenarios.

\begin{figure*}[!htb]
    \centering
    \setlength{\abovecaptionskip}{0pt}
    \setlength{\belowcaptionskip}{0pt}
    \subfigure[Test (COCO2017).]{
        \begin{minipage}[t]{0.30\linewidth}
            \centering
            \includegraphics[width=\linewidth]{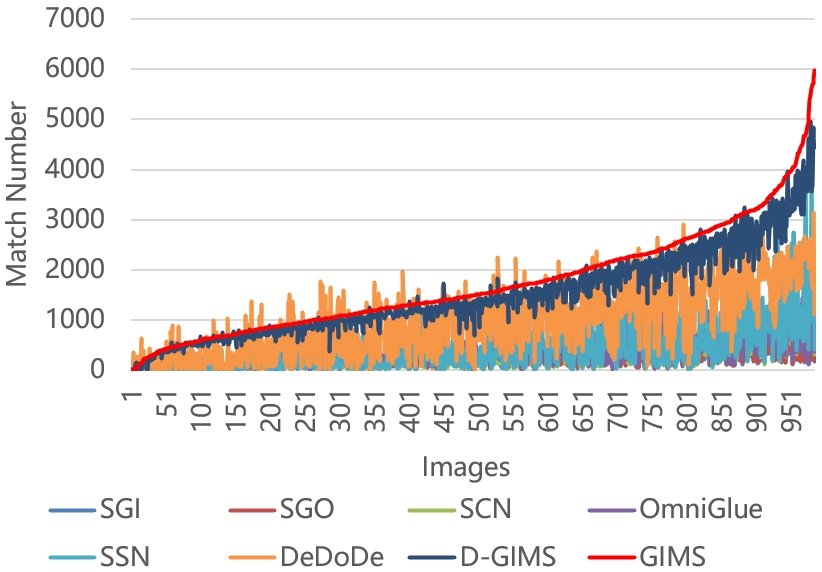}
        \end{minipage}
    }
    \subfigure[Indoor (RGB-D).]{
        \begin{minipage}[t]{0.30\linewidth}
            \centering
            \includegraphics[width=\linewidth]{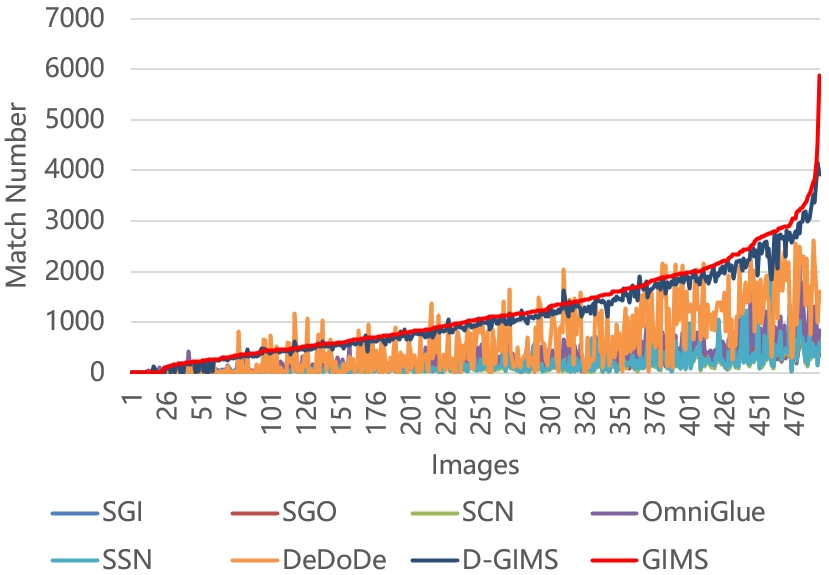}
        \end{minipage}
    }
    \subfigure[Outdoor (RGB-D).]{
         \begin{minipage}[t]{0.30\linewidth}
            \centering
            \includegraphics[width=\linewidth]{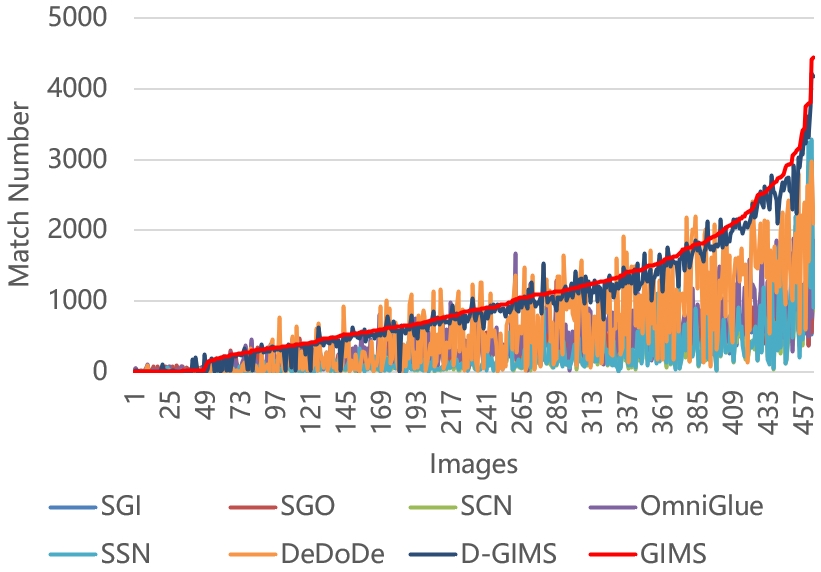}
        \end{minipage}
    }
    \caption{\textbf{Match Number of different methods.} In the legend, SGI represents SuperGlue with indoor model, SGO represents SuperGlue with outdoor model, SSN represents SIFT+SIFT+NNDR, SCN represents SIFT+CAR-HyNet+NNDR, and D-GIMS represents GIMS using the Delaunay graph construction method. We evaluate on three datasets and our method outperforms others, achieving an average improvement of up to 40.3 times overall.}
    \label{fig:valid_matching}
    \vspace{-1em}
\end{figure*}

\subsection{Evaluation of Valid Matching with Match Number}
The valid matching number is an intuitive performance metric that clearly reflects the performance of a matching method on a specific task, directly affecting the usability of the system. In Table~\ref{table:pose_estimation_auc}, we add the Average Match Number (AMN) for each method across three test sets. Additionally, Fig.~\ref{fig:valid_matching} further illustrates a detailed comparison of the match number among these methods. Note that we sort the results for clarity. The evaluation results show that different methods follow a consistent trend. Specifically, our method outperforms other methods by an average of 3.8x to 40.3x in the number of matches. This significant improvement is attributed to the ability of our method to efficiently identify and match similar image features in diverse scenarios, facilitated by the effective use of localized information derived from GNN. Moreover, the match number in GIMS remains slightly higher than in D-GIMS, further demonstrating the effectiveness of the proposed adaptive graph construction method.

\begin{figure*}[!t]
    \centering
    \setlength{\abovecaptionskip}{0em}
    \setlength{\belowcaptionskip}{0em}
    \includegraphics[width=\linewidth]{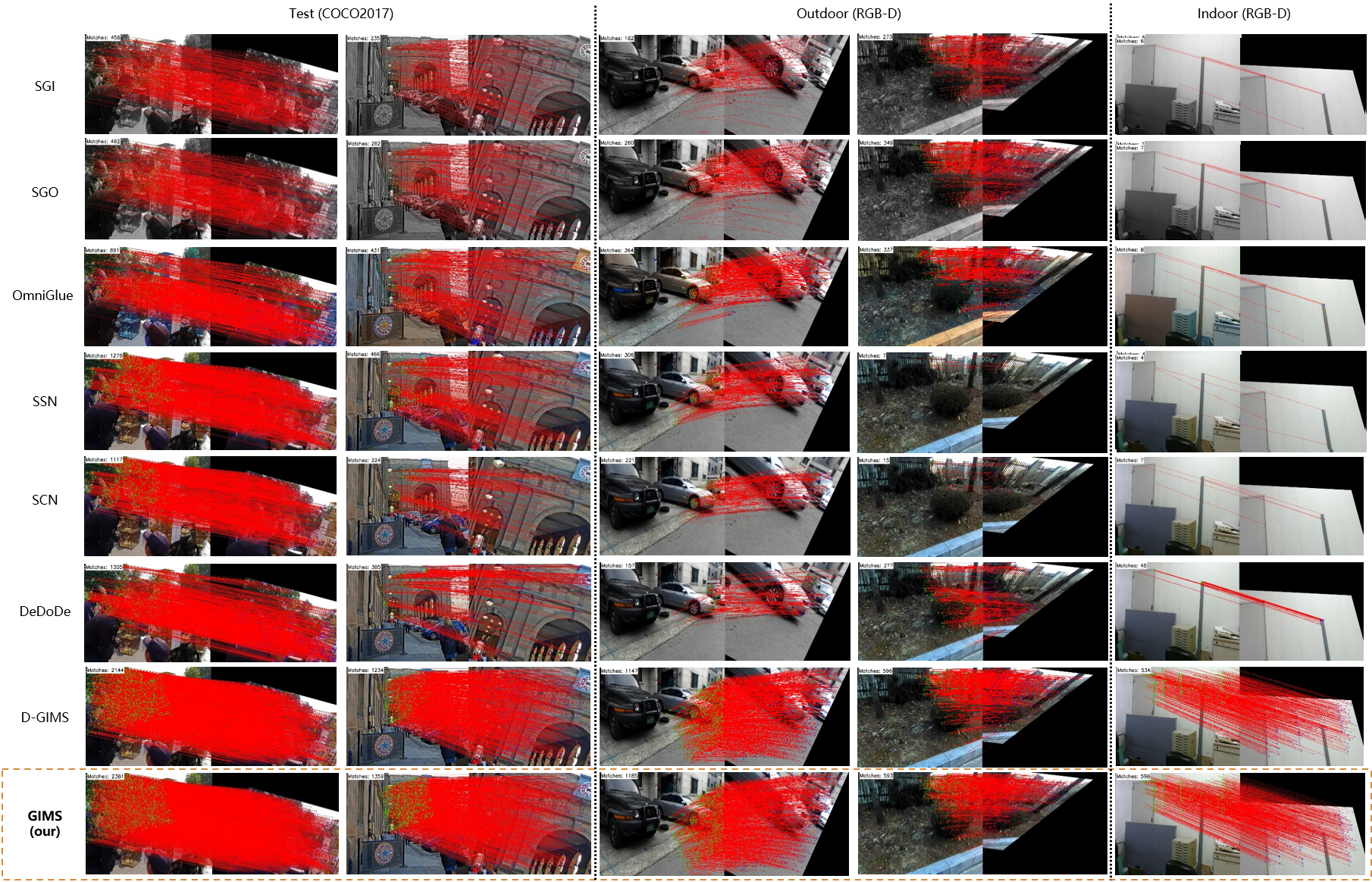}
    \caption{\textbf{Matching visualization of different methods.} We compare the matching results on selected images from three test sets. The red lines indicate the correct matches. Compared with other methods, GIMS produces more correct matches in all scenarios, even in challenging scenarios, reflecting its strong robustness and high recall.} 
    \label{fig:real_matches}
\end{figure*}

\begin{figure*}[!t]
    \centering
    \setlength{\abovecaptionskip}{0pt}
    \setlength{\belowcaptionskip}{0pt}
    \includegraphics[width=\linewidth]{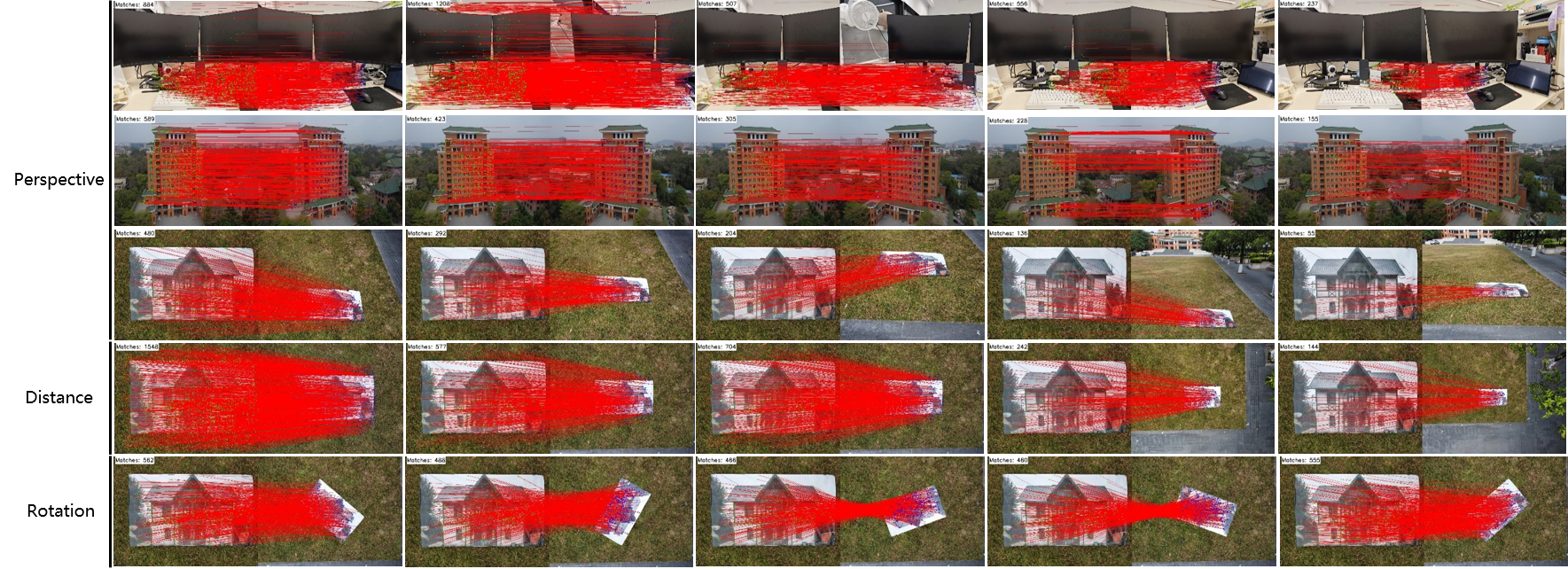}
    \caption{\textbf{Matching visualisation of GIMS on real-world images under varying conditions.} The red lines indicate the correct matches. Within each row, the difficulty increases from left to right. The consistent presence of correct matches across different transformations demonstrates the robustness of GIMS and its ability to handle diverse and challenging visual scenarios.}
    \label{fig:real_matches2}
     \vspace{-1em}
\end{figure*}

Additionally, we compare the actual matching results of different methods by selecting images from three test sets, as shown in Fig.~\ref{fig:real_matches}. The results indicate that, for images with minor changes, all methods achieve good matching performance. However, as difficulty increases, the matching performance of all methods decreases to varying degrees. Furthermore, DeDoDe tends to select regions with significant pixel changes, and the detected vertices are clustered together. This clustering makes the matching performance more susceptible to the influence of local regions. In contrast, our method significantly improves the match number in all cases, even in challenging scenarios. This is because the GNN model allows vertices to integrate information from neighbors, providing a more robust feature representation than a single vertex, even when the image undergoes transformations. Furthermore, self-attention and cross-attention enable vertices to understand their global positional information in both the current image and the image to be matched. However, due to space limitations, we present a subset of images here, while others exhibit the same trend. 

\subsection{Evaluation of Valid Matching in Oxford-Affine Dataset}
The Oxford Affine dataset~\citep{mikolajczyk2005comparison} is designed for benchmarking affine regions. It includes eight scenarios with variations in viewpoint, scale, rotation, and illumination. Each scenario consists of six image sequences with increasing levels of difficulty. This dataset is useful for evaluating algorithm performance in detecting and describing affine-invariant regions. 

The evaluation results are shown in Table~\ref{table:oxford_affine_dataset}. SGO and SGI have fewer matches in many scenarios and perform particularly poorly in the \textit{bark} scenario. OmniGlue is great at working with images that are rich in texture information such as \textit{leuven}, \textit{ubc} and \textit{wall}. SSN and SCN achieve relatively high matches in most scenarios. Additionally, SCN uses the CAR-HyNet descriptor, which has a stronger descriptive power compared to SIFT, explaining why SCN outperforms SSN in some scenarios. DeDoDe generally achieves good matching performance but is very unstable in certain scenarios. For example, it has fewer matches in the \textit{bark}, \textit{boat}, and \textit{graf} scenarios. This might be due to its lack of generality, poor adaptability to different scenes, and low feature detection capability in smooth regions and rotations. GIMS combines the classic SIFT detector and the advanced CAR-HyNet descriptor, leveraging the strengths of both. Its graph neural network-based matching method captures complex relationships between keypoints more effectively. Consequently, GIMS achieves significantly more matches in most scenarios, demonstrating a notable advantage and indicating its stability and superiority across different types of scenarios. Note that for D-GIMS, its matching performance is superior to other methods but slightly lower than that of GIMS. Considering that the only difference between D-GIMS and GIMS is the graph construction method, the reason for the difference is that the proposed AGC can provide vertices with better feature representations compared to Delaunay triangulation, which is consistent with previous experimental results.

\begin{table*}[!ht]
    \centering
    \setlength{\abovecaptionskip}{5pt}
    \setlength{\belowcaptionskip}{5pt}
    \caption{\textbf{Valid Matching in Oxford-Affine dataset.} In each scenario, the first image is matched with the other images, resulting in matching results for only five images being provided. GIMS outperformes others in most scenarios.}
    \label{table:oxford_affine_dataset}
    \resizebox{\linewidth}{!}{%
        \renewcommand{\arraystretch}{1.2}
        \begin{threeparttable}
            \begin{tabular}{lcccccccccccccccccccc}
                \Xhline{1.5pt}
                \multirow{2}{*}{\textbf{Label}} & \multicolumn{5}{c}{\textbf{bark}} & \multicolumn{5}{c}{\textbf{bikes}} & \multicolumn{5}{c}{\textbf{boat}} & \multicolumn{5}{c}{\textbf{graf}} \\
                \cmidrule(lr){2-6}\cmidrule(lr){7-11}\cmidrule(lr){12-16}\cmidrule(lr){17-21}
                 & \textbf{2} & \textbf{3} & \textbf{4} & \textbf{5} & \textbf{6} & \textbf{2} & \textbf{3} & \textbf{4} & \textbf{5} & \textbf{6} & \textbf{2} & \textbf{3} & \textbf{4} & \textbf{5} & \textbf{6} & \textbf{2} & \textbf{3} & \textbf{4} & \textbf{5} & \textbf{6} \\
                 \Xhline{1pt}
                SGO~\citep{sarlin2020superglue} & 332 & 0 & 4 & 94 & 0 & 668 & 601 & 409 & 244 & 143 & 689 & 620 & 0 & 207 & 207 & 588 & 435 & 341 & 237 & \textbf{198} \\
                SGI~\citep{sarlin2020superglue} & 319 & 4 & 6 & 0 & 0 & 666 & 598 & 394 & 241 & 141 & 690 & 606 & 5 & 160 & 0 & 576 & 421 & 303 & 203 & 49 \\
                OmniGlue~\citep{jiang2024Omniglue} & 292 & 6 & 7 & 71 & 8 & 1430 & 976 & 418 & 194 & 95 & 1730 & 1425 & 8 & 454 & 139 & 1136 & 919 & 778 & \textbf{548} & 103 \\
                DeDoDe~\citep{edstedt2024dedode} & 18 & 5 & 0 & 6 & 0 & 1715 & 1294 & 692 & 308 & 176 & 2313 & 301 & 8 & 160 & 0 & 1319 & 824 & 35 & 368 & 8 \\
                SSN~\citep{lowe1999object} & 611 & 567 & 680 & 451 & \textbf{259} & 767 & 564 & 301 & 226 & 153 & 2514 & 1887 & 661 & 481 & 201 & 1069 & 464 & 97 & 8 & 8 \\
                SCN~\citep{song2023carhynet} & 588 & 519 & 524 & 519 & 155 & 836 & 619 & 322 & 240 & 151 & 3047 & 2260 & 838 & 550 & 184 & 960 & 593 & 198 & 30 & 6 \\
                D-GIMS$^*$ & 2312 & 851 & 693 & 479 & 10 & 3264 & 2608 & 1418 & 1048 & 804 & 3556 & 2651 & 1024 & 657 & 256 & 2860 & 1758 & 919 & 267 & 14\\
                \textbf{GIMS$^*$} & \textbf{2503} & \textbf{910} & \textbf{783} & \textbf{540} & 9 & \textbf{3485} & \textbf{2705} & \textbf{1458} & \textbf{1065} & \textbf{815} & \textbf{4024} & \textbf{2841} & \textbf{1081} & \textbf{752} & \textbf{294} & \textbf{2983} & \textbf{1902} & \textbf{939} & 275 & 66 \\
                \midrule\specialrule{0em}{0.5pt}{0.5pt}\midrule
                \multirow{2}{*}{\textbf{Label}} & \multicolumn{5}{c}{\textbf{leuven}} & \multicolumn{5}{c}{\textbf{trees}} & \multicolumn{5}{c}{\textbf{ubc}} & \multicolumn{5}{c}{\textbf{wall}} \\
                 \cmidrule(lr){2-6}\cmidrule(lr){7-11}\cmidrule(lr){12-16}\cmidrule(lr){17-21}
                 & \textbf{2} & \textbf{3} & \textbf{4} & \textbf{5} & \textbf{6} & \textbf{2} & \textbf{3} & \textbf{4} & \textbf{5} & \textbf{6} & \textbf{2} & \textbf{3} & \textbf{4} & \textbf{5} & \textbf{6} & \textbf{2} & \textbf{3} & \textbf{4} & \textbf{5} & \textbf{6} \\
                 \Xhline{1pt}
                SGO~\citep{sarlin2020superglue} & 872 & 850 & 822 & 757 & 705 & 412 & 372 & 277 & 180 & 100 & 666 & 614 & 483 & 300 & 201 & 722 & 656 & 534 & 446 & 310 \\
                SGI~\citep{sarlin2020superglue} & 872 & 849 & 821 & 756 & 702 & 408 & 377 & 271 & 168 & 86 & 665 & 613 & 479 & 295 & 182 & 718 & 637 & 501 & 394 & 195 \\
                OmniGlue~\citep{jiang2024Omniglue} & 2006 & 1941 & 1856 & 1633 & 1301 & 1110 & 1060 & 682 & 377 & 155 & 1277 & 1118 & 1106 & 854 & 808 & 1723 & 1471 & 1225 & 1027 & 592 \\
                DeDoDe~\citep{edstedt2024dedode} & 3125 & 2940 & 2390 & 2476 & 1989 & 1749 & 1526 & 860 & 396 & 93 & 3358 & 2770 & 2340 & 1678 & \textbf{965} & 3241 & 3101 & 2405 & \textbf{1986} & \textbf{922} \\
                SSN~\citep{lowe1999object} & 1142 & 897 & 699 & 590 & 393 & 1906 & 1672 & 628 & 282 & 127 & 3150 & 2543 & 1609 & 745 & 327 & \textbf{5531} & 4142 & 2496 & 737 & 27 \\
                SCN~\citep{song2023carhynet} & 1191 & 952 & 761 & 632 & 407 & 2746 & \textbf{2675} & 1136 & 505 & 172 & 3359 & 2789 & 2100 & 903 & 431 & 5394 & 4502 & 3114 & 1456 & 244 \\
                D-GIMS$^*$ & 3782 & 3472 & 3253 & 3009 & 2697 & 2609 & 2159 & 1448 & 714 & 323 & 4314 & 3667 & 2940 & 1581 & 818 & 5196 & 4561 & 3258 & 1798 & 15 \\
                \textbf{GIMS$^*$} & \textbf{4056} & \textbf{3714} & \textbf{3417} & \textbf{3137} & \textbf{2798} & \textbf{2779} & 2287 & \textbf{1566} & \textbf{772} & \textbf{380} & \textbf{4742} & \textbf{3998} & \textbf{3172} & \textbf{1718} & 881 & 5485 & \textbf{4759} & \textbf{3433} & 1901 & 396 \\
                \Xhline{1.5pt}
            \end{tabular}
            \begin{tablenotes}
              \item $^*$This Work.
            \end{tablenotes}
        \end{threeparttable}
    }
    \vspace{-1em}
\end{table*}

\subsection{Evaluation of Valid Matching in Real-world}
To evaluate the matching performance of the proposed system in real-world scenarios, we conduct experiments using several sets of images taken with drones and phones. These images are categorized into three groups: Perspective, Distance, and Rotation. Each group presents the matching results for five pairs of images. Since the transformation matrices of these images are unknown, we cannot compute the AUC. Therefore, we used the match number as the evaluation metric. The evaluation results are shown in Fig.~\ref{fig:real_matches2}.

The perspective transformation typically involves changes in the camera viewpoint, leading to significant changes in the appearance of objects in the images. Despite the high complexity of image changes caused by perspective transformations, GIMS still finds a large number of correct matches, demonstrating its robustness in handling viewpoint changes. The distance variation involves changes in the distance between the camera and the object, resulting in changes in the size of the object in the images. GIMS shows varying performance in handling distance changes, as scale changes significantly impact the matching performance. However, GIMS exhibits a degree of scale invariance. The rotation transformation involves the rotation of objects while maintaining their shape and proportions. Under rotational transformations, GIMS performs very reliably, with all image pairs having more than 400 matches. This indicates that GIMS is highly robust in finding and matching vertices when handling rotations. Note that GIMS uses SIFT to detect vertices and construct the graph. Since SIFT excels at extracting key points in areas with distinct edges, corners, or rich textures, there are fewer vertices in smooth regions such as the sky, grass, or monitors. Overall, the results indicate that GIMS can adapt to different scenarios and perform well even under extreme conditions, demonstrating strong robustness and matching capability.

\subsection{Latency Analysis}
In Section \ref{sec:complexity}, we analyze the computational complexity of each stage in GIMS from a theoretical perspective. To further provide an intuitive view of the actual runtime cost and the contribution of each stage to the overall efficiency, we introduce a detailed runtime latency analysis in this section. Specifically, we measure the time consumed by key stages in the system and compare GIMS with several representative methods, as shown in Table~\ref{tab:latency_analysis}. Inference is performed on an NVIDIA RTX 3090 GPU (24 GB).

\begin{table*}[!htbp]
    \belowrulesep=0pt
    \aboverulesep=0pt
    \renewcommand{\arraystretch}{1.1}
    \centering
    \caption{Comparison of runtime latency, correct matches, and memory usage across methods. KD=Keypoint Detection, PG=Patch Generation, DG=Descriptor Generation, GC=Graph Construction, KP=Number of Keypoints.}
    \label{tab:latency_analysis}
    \resizebox{0.95\linewidth}{!}{%
        \begin{tabular}{lcccccccc|ccc||cccc}
            \Xhline{1.5pt}
            \multicolumn{1}{c}{\multirow{3}{*}{\textbf{Method}}} & \multicolumn{11}{c||}{\textbf{Latency (s)}} & \multicolumn{4}{c}{\textbf{Statistics}} \\
            \cmidrule(lr){2-12} \cmidrule(lr){13-16}
            & \multicolumn{4}{c}{\textbf{Image 1}} & \multicolumn{4}{c|}{\textbf{Image 2}} & \multirow{2}{*}{\textbf{Matching}} & \multirow{2}{*}{\textbf{RANSAC}} & \multirow{2}{*}{\textbf{Total}} & \multirow{2}{*}{\textbf{KP1}} & \multirow{2}{*}{\textbf{KP2}} & \multirow{2}{*}{\textbf{\begin{tabular}[c]{@{}c@{}}Correct\\ Matches\end{tabular}}} & \multirow{2}{*}{\textbf{\begin{tabular}[c]{@{}c@{}}Peak\\Memory\end{tabular}}} \\
            \cmidrule(lr){2-5} \cmidrule(lr){6-9}
            & \textbf{KD} & \textbf{PG} & \textbf{DG} & \textbf{GC} & \textbf{KD} & \textbf{PG} & \textbf{DG} & \textbf{GC} & & & & & & & \\
            \Xhline{1pt}
            \textbf{SGO}      & 0.143 & --   & 0.035 & --   & 0.009 & --   & 0.001 & --   & 0.531 & 0.002 & 0.800  & 1455  & 1235  & 743  & 0.49 GB \\
            \textbf{SGI}      & 0.146 & --   & 0.037 & --   & 0.007 & --   & 0.001 & --   & 0.486 & 0.006 & 0.737  & 1455  & 1235  & 716  & 0.49 GB \\
            \textbf{OmniGlue} & 2.825 & --   & 4.536 & --   & 0.767 & --   & 3.764 & --   & 7.404 & 0.001 & 19.367 & 2968  & 2449  & 1501 & 22.20 GB \\
            \textbf{DeDoDe}   & 0.579 & --   & 0.069 & --   & 0.051 & --   & 0.042 & --   & 0.101 & 0.001 & 4.963  & 10000 & 10000 & 239  & 1.65 GB \\
            \textbf{SSN}      & 0.205 & --   & 0.328 & --   & 0.137 & --   & 0.259 & --   & 1.507 & 0.002 & 2.478  & 8849  & 6558  & 1887 & 0.14 GB \\
            \textbf{SCN}      & 0.203 & 1.935 & 0.486 & --   & 0.139 & 1.448 & 0.210 & --   & 1.753 & 0.002 & 6.358  & 8849  & 6558  & 2260 & 0.81 GB \\
            \textbf{D-GIMS}   & 0.181 & 2.306 & 0.528 & 1.199 & 0.165 & 2.160 & 0.308 & 0.886 & 1.744 & 0.041 & 9.625  & 10000 & 10000 & 2604 & 3.23 GB \\
            \textbf{GIMS}     & 0.172 & 2.238 & 0.526 & 2.994 & 0.186 & 2.266 & 0.304 & 2.651 & 1.649 & 0.012 & 13.119 & 10000 & 10000 & \textbf{2808} & 3.21 GB \\
            \Xhline{1.5pt}
        \end{tabular}%
    }
\end{table*}

The experimental results show that GIMS incurs relatively high latency in the patch generation and graph construction stages, which is consistent with our expectations. Nevertheless, GIMS still achieves the best performance, with the number of correct matches significantly exceeding that of other methods. In terms of resource consumption, under the condition of 20,000 vertices, the peak memory usage of GIMS is 3.21 GB, which is substantially lower than that of the OmniGlue, indicating that GIMS maintains high matching performance while offering good deployability and scalability. These results also provide clear guidance for future optimization directions of the system, such as compressing the graph construction process and the feature generation module to improve overall runtime efficiency.

\subsection{Ablation Study}

\subsubsection{Impact of GNN Layers}
To explore the impact of GNN layer depth on matching performance, we evaluate the performance of GraphSAGE model with varying numbers of layers on the pose estimation task, training them for the same number of epochs on the COCO2017 dataset. We provide the AUC for models with different layers at thresholds of 5, 10, and 25. Here, \textit{Base} represents the baseline model without a GNN.

\begin{table}[!htbp]
    \vspace{-0.5em}
    \setlength{\abovecaptionskip}{0pt}
    \setlength{\belowcaptionskip}{0pt}
    \centering
    \caption{\textbf{Performance of different layers of GraphSAGE}. \textit{Base} represents the baseline without the GNN model. The model achieves excellent performance with just 3 layers.}
    \label{tab:gnn_layers}
    \setlength\tabcolsep{20pt}
    \renewcommand{\arraystretch}{1.2}
    \resizebox{0.9\columnwidth}{!}{%
        \begin{tabular}{lcccc}
            \Xhline{1.5pt}
            \noalign{\vskip 1pt}
            \multirow{2}{*}{\textbf{Layers}} & \multicolumn{3}{c}{\textbf{Pose Estimation AUC}}\\ 
            \noalign{\vskip 1pt}
            \cline{2-4}
            \noalign{\vskip 1pt}
            & \textbf{@5} & \textbf{@10} & \textbf{@25} \\ 
            \Xhline{1pt}
            Base             & 66.48	      & 77.93	        & 87.92           \\
            1-Layer          & 76.59          & 86.08           & 93.09           \\
            2-Layer          & 76.44          & 86.21           & 93.17           \\
            \textbf{3-Layer} & \textbf{78.63} & \textbf{87.74}	& \textbf{93.92}  \\
            4-Layer          & 77.54	      & 86.89	        & 93.63           \\
            5-Layer          & 77.00	      & 86.66	        & 93.58           \\
            \Xhline{1.5pt}
        \end{tabular}%
    }
    \vspace{-0.5em}
\end{table}

Table~\ref{tab:gnn_layers} shows that compared to the \textit{Base}, the 1-layer GraphSAGE model significantly improves the AUC across all three metrics, indicating that even just one GNN layer can substantially improve model performance. The AUC further improves with the 2-layer and 3-layer models. However, we also observe that when the number of GNN layers increases beyond three, the AUC starts to decline. This suggests that adding more layers does not lead to better performance, instead resulting in diminishing returns compared to the 3-layer model. The optimal number of layers allows the model to capture sufficient neighborhood information, thereby improving recognition and classification accuracy. As the number of layers increases, the aggregation of information from distant neighbors causes the feature representations of different vertices to become similar, making it difficult for the model to distinguish between them. This phenomenon, known as over-smoothing, diminishes the discriminative power of vertex features and negatively impacts the model's classification performance. Additionally, using extra layers requires more computational resources and increases latency.

Considering that the 3-layer GraphSAGE model performs best across all three metrics while avoiding computational redundancy, we chose a 3-layer design for GIMS.

\subsubsection{Ablation Analysis of GIMS}

The proposed GIMS framework comprises four key components: the AGC module for building the graph structure, the SIFT keypoint detector, the CAR-HyNet descriptor for feature representation, and the GNNMatcher module for graph matching. To evaluate the contribution of each component, we design a progressive ablation strategy reflected in four representative system variants: SSN, SCN, D-GIMS, and GIMS. The configuration of each variant is summarized in Table~\ref{tab:ablation_gims}. It is important to note that these four configurations are already included in the main experiments; here, they are reorganized to provide a structured ablation analysis.

Specifically, SSN serves as the baseline configuration with handcrafted descriptors and traditional nearest-neighbor matching. SCN replaces the descriptor with CAR-HyNet to evaluate the impact of learned features. D-GIMS introduces the GNNMatcher on top of a Delaunay graph to assess the benefit of graph-based matching. Finally, GIMS integrates the full pipeline with AGC, allowing a direct comparison between adaptive and heuristic graph construction strategies. 

\begin{table}[h]
    \centering
    \caption{Configurations of system variants used in the ablation analysis of GIMS.}
    \label{tab:ablation_gims}
    \resizebox{\linewidth}{!}{%
        \begin{tabular}{lccccc}
            \toprule
            \textbf{Stage} & \textbf{Graph} & \textbf{Detector} & \textbf{Descriptor} & \textbf{Matcher}  & \textbf{Label in Exp.} \\
            \midrule
            \textbf{Baseline}    & None       & SIFT   & SIFT       & NNDR & SSN       \\
            \textbf{+ CAR-HyNet} & None       & SIFT   & CAR-HyNet  & NNDR & SCN       \\
            \textbf{+ GNNMatcher}  & Delaunay   & SIFT   & CAR-HyNet  & GNNMatcher & D-GIMS \\
            \textbf{+ AGC}    & AGC        & SIFT   & CAR-HyNet  & GNNMatcher & GIMS \\
            \bottomrule
        \end{tabular}%
    }
\end{table}

Table~\ref{table:pose_estimation_auc} reports the AUC for pose estimation, while Figures~\ref{fig:valid_matching} and~\ref{fig:real_matches} and Table~\ref{table:oxford_affine_dataset} present the match number. Experimental results show that SSN detects more correct matches than SCN, although it exhibits slightly lower performance in terms of AUC. By introducing explicit graph construction and a GNN-based matcher, D-GIMS significantly increases the number of correct matches, and also brings a noticeable improvement in AUC. Finally, GIMS further boosts both AMN and AUC by leveraging the proposed AGC, demonstrating the effectiveness of adaptive topological modeling in enhancing both matching quantity and geometric consistency. These ablation results collectively validate the effectiveness and necessity of each key component in our proposed system.

\subsection{Multi-GPU Parallel Training Acceleration}
A large number of keypoints pose significant challenges to model training. Implementing a multi-GPU parallel training strategy can significantly improve training efficiency. Here, we conduct a preliminary exploration of the impact of different numbers of GPUs on training efficiency to provide guidance and support for subsequent model optimization and large-scale data training. We measure the time taken to complete a single epoch of training on the COCO2017 dataset using different numbers of GPUs on a single machine. Given that training with 10,000 keypoints requires more than 35 GB of GPU memory, and the RTX 3090 is limited to 24 GB, we limit each image to 2,048 keypoints, totaling 4,096 keypoints. For scenarios involving fewer than 2,048 keypoints, we randomly selected positions on the image to meet the target number of keypoints. The experimental setup was based on the GPU groupings detailed in Table~\ref{tab:gpu_grouping}, with the test results illustrated in Fig.~\ref{fig:training_time}.

\begin{table}[!htbp]
    \vspace{-1em}
    \centering
    \setlength{\abovecaptionskip}{0pt}
    \setlength{\belowcaptionskip}{0pt}
    \caption{\textbf{Grouping of 2 RTX 3090 and 2 Tesla A40 GPUs for parallel training.} The $\checkmark$ indicates that the GPU is used in the group, while the $\times$ indicates that it is not used in the group.}
    \label{tab:gpu_grouping}
    \setlength\tabcolsep{6pt}
    \resizebox{0.9\columnwidth}{!}{%
        \renewcommand{\arraystretch}{1.1}
        \begin{tabular}{ccccc}
            \Xhline{1.5pt}
            \noalign{\vskip 1pt}
            \textbf{Group No.} & \textbf{\makecell[c]{RTX 3090 \\ (ID 1)}} & \textbf{\makecell[c]{RTX 3090 \\ (ID 2)}} & \textbf{\makecell[c]{Tesla A40 \\ (ID 3)}} & \textbf{\textbf{\makecell[c]{Tesla A40 \\ (ID 4)}}} \\
            \noalign{\vskip 1pt}
            \Xhline{1pt}
            \noalign{\vskip 1pt}
            1 & $\times$      &  $\times$     &  $\times$     & $\times$      \\
            2 & $\checkmark$  &  $\times$     &  $\times$     & $\times$      \\
            3 & $\times$      &  $\times$     &  $\checkmark$ & $\times$      \\
            4 & $\checkmark$  &  $\checkmark$ &  $\times$     & $\times$      \\
            5 & $\times$      &  $\times$     &  $\checkmark$ & $\checkmark$  \\
            6 & $\checkmark$  &  $\checkmark$ &  $\checkmark$ & $\times$      \\
            7 & $\checkmark$  &  $\times$     &  $\checkmark$ & $\checkmark$  \\
            8 & $\checkmark$  &  $\checkmark$ &  $\checkmark$ & $\checkmark$  \\
            \Xhline{1.5pt}
        \end{tabular}%
    }
\end{table}

\begin{figure}[!htb]
    \centering
    \setlength{\abovecaptionskip}{0em}
    \setlength{\belowcaptionskip}{0em}
    \includegraphics[width=0.9\linewidth]{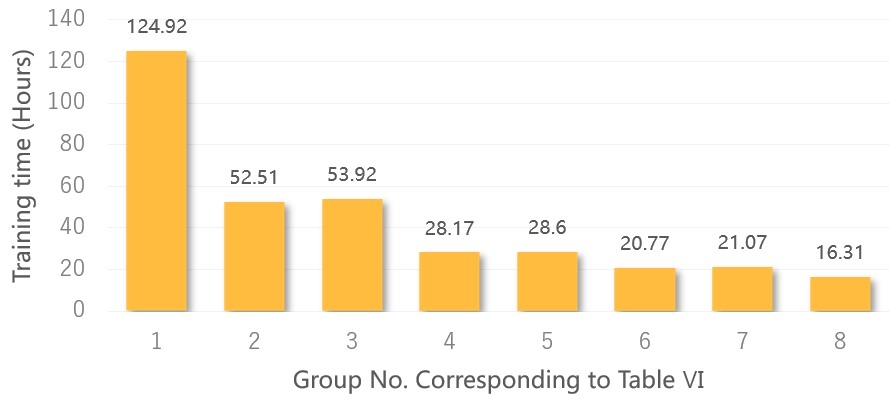}
    \caption{\textbf{Training time of different GPU groups on the COCO2017 dataset.} The grouping details are shown in Table~\ref{tab:gpu_grouping}. The training time decreases as the number of GPUs increases; however, this decrease is not linear.}
    \label{fig:training_time}
    \vspace{-1em}
\end{figure}

The experimental results show that, compared to CPUs, using GPUs significantly reduces training time. For the same amount of data, the RTX 3090 takes slightly less training time than the A40, indicating that with sufficient memory, the overall performance of the RTX 3090 is marginally better. Furthermore, the overall training time tends to decrease as the number of GPUs involved in training increases. However, we also find that the reduction in time is not proportional to the number of GPUs. This is mainly because, under the multi-GPU data parallel training framework, each GPU needs to synchronize gradient information after each iteration, a process that involves significant communication overhead. As the number of GPUs involved in the computation increases, this communication cost grows, which in turn affects the overall training efficiency. Moreover, while most of the deep learning computational burden is borne by the GPUs, tasks such as data preprocessing, loading, and its transfer to the GPUs are performed by the CPUs. Therefore, an increase in the number of GPUs leads to a corresponding increase in the demand for CPU processing power, affecting the training time.

\section{Discussion and Future Work}
\label{sec:discussion}
The proposed system shows excellent image matching performance. However, as a preliminary study, we also notice that there are several areas for improvement. We share our thoughts below for discussion.

\textbf{More appropriate GNN models.} We believe there are many advanced GNN models that are better suited for image matching. In future work, we will explore these models and conduct in-depth studies to find the optimal solutions.
    
\textbf{Faster graph construction methods.} Accurate graph construction methods are time-consuming. Appropriate graph construction methods can enhance efficiency, such as using more appropriate data structures and faster neighborhood search algorithms. We will continue to optimize the current graph construction method so that it can be faster.

\textbf{Efficient optimal transport methods.} In this work, we use the conventional Sinkhorn algorithm to solve the assignment matrix. However, there are superior algorithms that can more efficiently determine the final assignment of features and produce optimal matching results, such as Dual-Softmax~\citep{cheng2021improving}. In future work, we will consider incorporating Dual-Softmax into the proposed system to further enhance performance.

\textbf{More appropriate keypoint detection methods}. High-quality keypoints are crucial for graph construction and also have a significant impact on subsequent matching tasks. In this work, we use only SIFT for keypoint detection. In fact, there are many other keypoint detection methods and feature extraction techniques with better performance. These methods can improve detection accuracy and efficiency, thus improving the effectiveness of subsequent matching tasks. We plan to explore and apply these advanced techniques in our future work to further enhance overall performance.

\textbf{Parallel and distributed training}. As the graph size increases, memory and computation usage also increase. Parallel and distributed training helps in handling larger datasets and significantly reduces training time. However, efficiently conducting parallel and distributed training for graph neural networks remains challenging. We have started related research and will continue to study and optimize these methods to achieve higher training efficiency.


\section{Conclusion}
\label{sec:conclusion}
In this paper, we propose GIMS, a novel image matching system based on a similarity-aware adaptive graph construction method and a graph neural network-based image matching method. The proposed graph construction method adapts GNNs to images by establishing edges between vertices based on neighborhood distance and feature similarity, and dynamically adjusting criteria for incorporating new vertices according to existing vertex features. This approach constructs precise and robust graph structures, avoiding redundant vertices and edges. Furthermore, we use GNN to explicitly learn local feature encodings of vertices, followed by learning the spatial and global feature encoding of vertices through positional encoding and attentional Transformer. This method helps to improve the local and global representation of vertices in the graph structure. Finally, we employ the Sinkhorn algorithm to iteratively determine the optimal matching results. We train our system using multiple GPUs on a single machine. We then perform experiments on large image datasets and real-world images. Our experiments show that, compared to existing methods, our system achieves significant improvements in image matching performance, where the overall matching results achieve an improvement of 3.8 to as high as 40.3 times in commonly used benchmark datasets. We acknowledge that the current graph construction method is complex and time-consuming. In future work, we plan to explore optimization efforts, including more appropriate GNN models, faster graph construction techniques, efficient optimal transport methods, as well as more appropriate keypoint detection methods.

\section*{Acknowledgments}
\label{sec:acknowledgment}

Funding: This work was supported by the South China University of Technology Research Start-up Fund [grant numbers K3200890].

\bibliographystyle{elsarticle-harv} 
\bibliography{tex/references}
\end{document}